\documentclass[nonacm, sigconf]{acmart}

\usepackage{enumitem}
\usepackage{pifont}
\usepackage{placeins}
\usepackage{amsmath}
\usepackage{mathtools}
\usepackage{subcaption}
\usepackage{lipsum}
\usepackage{wasysym}
\usepackage{multirow}
\usepackage{tabularray}
\usepackage{graphicx}
\usepackage{bm}
\usepackage{newtxtext}
\usepackage[skip=5pt]{caption}

\AtBeginDocument{%
  \providecommand\BibTeX{{%
    \normalfont B\kern-0.5em{\scshape i\kern-0.25em b}\kern-0.8em\TeX}}}

\acmISBN{978-1-4503-XXXX-X/2018/06}

\begin{document}

\title{DQE: A Semantic-Aware Evaluation Metric for Time Series Anomaly Detection}

\author{Yuewei Li}
\affiliation{%
  \institution{Hangzhou Dianzi University}
  \city{Hangzhou}
  \country{China}
}

\author{Dalin Zhang}
\authornote{Corresponding Author: dalinz@cs.aau.dk}

\affiliation{
  \institution{Hangzhou Dianzi University}
  \city{Hangzhou}
  \country{China}
}

\author{Huan Li}
\affiliation{
  \institution{Zhejiang University}
  \city{Hangzhou}
  \country{China}
}

\author{Xinyi Gong}
\affiliation{
  \institution{Hangzhou Dianzi University}
  \city{Hangzhou}
  \country{China}
}

\author{Hongjun Chu}
\affiliation{
  \institution{Hangzhou Dianzi University}
  \city{Hangzhou}
  \country{China}
}

\author{Zhaohui Song}
\affiliation{
  \institution{Hangzhou Dianzi University}
  \city{Hangzhou}
  \country{China}
}

\begin{abstract}
Time series anomaly detection has achieved remarkable progress in recent years. However, evaluation practices have received comparatively less attention, despite their critical importance. Existing metrics exhibit several limitations: (1) bias toward point-level coverage, (2) insensitivity or inconsistency in near-miss detections, (3) inadequate penalization of false alarms, and (4) inconsistency caused by threshold or threshold-interval selection. These limitations can produce unreliable or counterintuitive results, hindering objective progress. In this work,
we revisit the evaluation of time series anomaly detection
from the perspective of detection semantics and propose a novel metric for more comprehensive assessment.
We first introduce a partitioning strategy grounded in detection semantics, which decomposes the local temporal region of each anomaly into three functionally distinct subregions.
Using this partitioning, we evaluate overall detection behavior across events and design finer-grained scoring mechanisms for each subregion, enabling more reliable and interpretable assessment.
Through a systematic study of existing metrics, we identify an evaluation bias associated with threshold-interval selection and adopt an approach that aggregates detection qualities across the full threshold spectrum, thereby eliminating evaluation inconsistency.
Extensive experiments on synthetic and real-world data demonstrate that our metric provides stable, discriminative, and interpretable evaluation, while achieving robust assessment compared with ten widely used metrics.

\noindent Public source code: \url{https://github.com/Yueweilirepo/DQE}
\end{abstract}


\maketitle

\section{Introduction}\label{sec:Introduction}

Time series anomaly detection (TSAD), which aims to identify abnormal temporal patterns amid normal behaviors, is crucial in diverse domains such as finance~\cite{song2023anomaly,bustamante2024financial}, cybersecurity~\cite{nakamura2025cybercscope,rookard2024unsupervised}, and the industrial Internet of Things ~\cite{gao2025dynamic,chen2025privacy,kaufman2025time}. This wide applicability has driven the development of numerous detection models~\cite{zhong2025multi,wu2024catch,li2025tsinr,huang2025graph,jang2025tail,liu2025gcad}. Equally important, however, is the advancement of fair and reliable evaluation metrics. An effective metric not only compares and ranks competing models but also guides research directions, validates methodological progress, and ensures that improvements translate into real-world utility.

\begin{table*}[htbp]
    \caption{Comparison of metrics on event-level coverage.
    Left: different detection results (P1-P2) against Ground Truth (GT).
    Right: corresponding evaluation scores of various metrics.
    Original-F denotes the standard F1-score.
    }
    \label{tab:bias_point_level_coverage}
    \begin{subtable}[t]{\textwidth}
        \label{tab:bias_point_level_coverage_data}
        \resizebox{\textwidth}{!}{
            \begin{tblr}{
                colspec = {ccccccccccccc},
                rowspec = {Q[m]Q[m]Q[m]Q[m]Q[m]Q[m]Q[m]},
                hline{1,Z} = {1pt},
                hline{3} = {3-Z}{0.75pt},
                hline{2} = {3-9}{0.5pt,leftpos = -1, rightpos = -1, endpos},
                hline{2} = {10-Z}{0.5pt,leftpos = -1, rightpos = -1, endpos},
                cell{1}{1} = {r=2}{},
                cell{1}{2} = {r=2}{},
                cell{1}{3} = {c=7}{},
                cell{1}{10} = {c=4}{},
                cell{1}{2-Z} = {font=\huge\bfseries},
                cell{2}{2-Z} = {font=\huge\bfseries},
                cell{1-Z}{1} = {font=\fontsize{17pt}{0pt}\selectfont\bfseries},
                cell{3-Z}{3-Z} = {font=\Huge},
                cell{3}{3-9} = {font=\Huge\bfseries},
                cell{4}{10-Z} = {font=\Huge\bfseries},
                cell{3-Z}{1} = {cmd=\raisebox{0.4cm}},
                cell{3-Z}{3-Z} = {cmd=\raisebox{0.4cm}},
            }

GT & \includegraphics{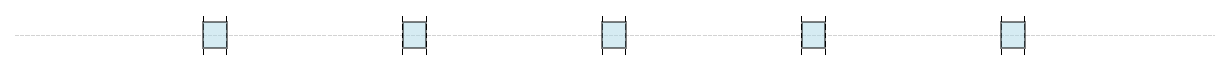} & w/ issue & AUC-ROC$^{\textcolor{cyan6}{*}}$ & AUC-PR$^{\textcolor{cyan6}{*}}$ & PA-K$^{\textcolor{purple4}{*}}$ & VUS-ROC$^{*}$ & VUS-PR$^{*}$ & PATE$^{*}$ & w/o issue & eTaF$^{\star}$ & AF$^{\star}$ & DQE$^{\star}$ (ours) \\

GT & \includegraphics{figures/single_prediction_figures/Event_Detection_Rate_full_detection_gt} & Original-F & AUC-ROC & AUC-PR & PA-K \cite{kim2022towards} & VUS-ROC \cite{paparrizos2022volume} & VUS-PR \cite{paparrizos2022volume} & PATE \cite{ghorbani2024pate} & RF \cite{tatbul2018precision} & eTaF \cite{hwang2022you} & AF \cite{huet2022local} & DQE (ours) \\

P1 & \includegraphics{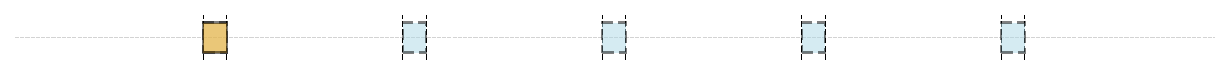} & 0.33 & 0.60 & 0.28 & 0.33 & 0.52 & 0.14 & 0.64 & 0.33 & 0.33 & 0.33 & 0.20 \\
P2 & \includegraphics{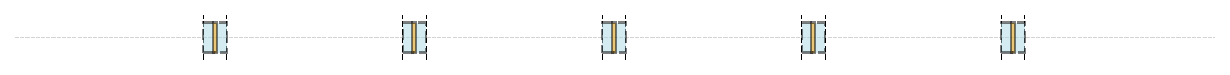} & 0.05 & 0.51 & 0.12 & 0.05 & 0.51 & 0.13 & 0.56 & 0.36 & 0.68 & 0.98 & 1.00 \\

            \end{tblr}
        }
    \end{subtable}
\end{table*}

\begin{table*}[htbp]
    \caption{Comparison of metric insensitivity in evaluating near-miss detections.}
    \label{tab:proximity_case}
    \begin{subtable}[t]{\textwidth}
        \label{tab:proximity_case_data}
        \resizebox{\textwidth}{!}{
            \begin{tblr}{
                colspec = {ccccccccccccc},
                rowspec = {Q[m]Q[m]Q[m]Q[m]Q[m]Q[m]Q[m]},
                hline{1,Z} = {1pt},
                hline{3} = {3-Z}{0.75pt},
                hline{2} = {3-8}{0.5pt,leftpos = -1, rightpos = -1, endpos},
                hline{2} = {9-Z}{0.5pt,leftpos = -1, rightpos = -1, endpos},
                cell{1}{1} = {r=2}{},
                cell{1}{2} = {r=2}{},
                cell{1}{3} = {c=6}{},
                cell{1}{9} = {c=5}{},
                cell{1}{2-Z} = {font=\huge\bfseries},
                cell{2}{2-Z} = {font=\huge\bfseries},
                cell{1-Z}{1} = {font=\fontsize{17pt}{0pt}\selectfont\bfseries},
                cell{3-Z}{3-Z} = {font=\Huge},
                cell{3}{9-Z} = {font=\Huge\bfseries},
                cell{3-Z}{1} = {cmd=\raisebox{0.4cm}},
                cell{3-Z}{3-Z} = {cmd=\raisebox{0.4cm}},
            }

GT & \includegraphics{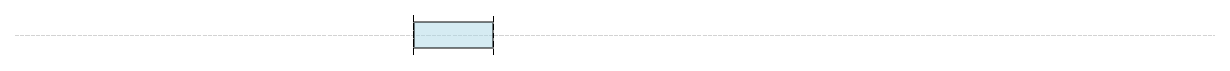} & w/ issue & VUS-PR$^{*}$ & PATE$^{*}$ & w/o issue & AUC-ROC$^{\textcolor{cyan6}{*}}$ & AUC-PR$^{\textcolor{cyan6}{*}}$ & w/o issue & RF$^{\star}$ & eTaF$^{\star}$ & AF$^{\star}$ & DQE$^{\star}$ \\

GT & \includegraphics{figures/single_prediction_figures/proximity_case_gt} & Original-F & AUC-ROC & AUC-PR & PA-K & RF & eTaF & VUS-ROC & VUS-PR & PATE & AF & DQE (ours) \\

P1 & \includegraphics{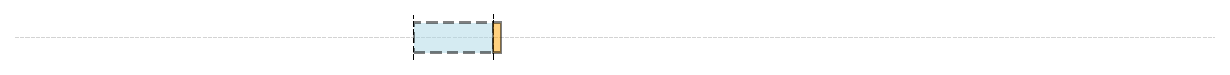} & 0.00 & 0.50 & 0.07 & 0.00 & 0.00 & 0.00 & 0.54 & 0.15 & 0.39 & 0.93 & 0.67 \\
P2 & \includegraphics{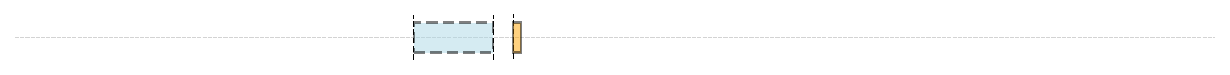} & 0.00 & 0.50 & 0.07 & 0.00 & 0.00 & 0.00 & 0.52 & 0.11 & 0.30 & 0.90 & 0.50 \\
P3 & \includegraphics{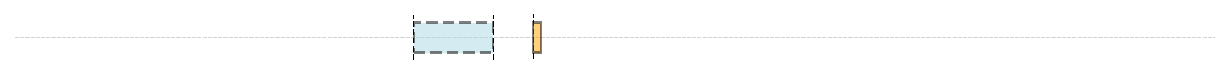} & 0.00 & 0.50 & 0.07 & 0.00 & 0.00 & 0.00 & 0.50 & 0.09 & 0.21 & 0.86 & 0.33 \\
P4 & \includegraphics{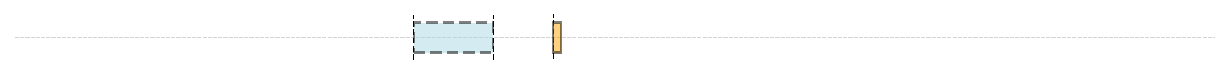} & 0.00 & 0.50 & 0.07 & 0.00 & 0.00 & 0.00 & 0.50 & 0.09 & 0.12 & 0.83 & 0.15 \\

            \end{tblr}
        }
    \end{subtable}
\end{table*}

Evaluating TSAD is challenging due to the ambiguous definition of successful detection in continuous temporal data.
Unlike classification tasks with clear instance-level labels, time series anomalies often lack clear boundaries.
Conventional metrics, which rely on counting True Positives (TPs) and False Positives (FPs), therefore
provide an incomplete assessment by enforcing a binary dichotomy on anomalies with ambiguous temporal boundaries.
This misalignment highlights the need for a comprehensive evaluation that considers both \textit{whether} anomalies are detected and \textit{how well} they are detected within their temporal context.
From this perspective, we identify several limitations of existing metrics.
(a) \textit{Bias toward point-level coverage (L1).} Many metrics emphasize the proportion of correctly detected points rather than the coverage of anomaly events, which are continuous temporal intervals
representing holistic semantic occurrences.
This bias distorts evaluation by favoring models that cover many points of a single anomaly event while entirely missing other distinct, semantically meaningful events (see \autoref{tab:bias_point_level_coverage}), causing valid detections to be substantially under-evaluated.
(b) \textit{Insensitivity or inconsistency in near-miss detections (L2).} Owing to temporal correlations, detections often occur near anomaly boundaries, exhibiting proximal detection behavior and providing valuable positional information.
However, existing metrics either disregard this behavior (\autoref{tab:proximity_case}) or evaluate it inconsistently as detections become less precise (\autoref{tab:inconsistency_in_near_section}, \autoref{tab:af_problem}), hindering fine-grained assessment of near-miss detections.
(c) \textit{Inadequate penalization of false alarms (L3).} Detections occurring far from any anomaly event represent spurious detection behavior, which can trigger unnecessary interventions and erode user trust.
Yet, many metrics penalize them insufficiently, assigning identical scores for substantially different false alarm patterns (\autoref{tab:false_alarm_only_num}).
Some metrics even assign high scores to random detections (\autoref{tab:random_case}), diminishing their ability to distinguish truly effective methods from poor ones.
These limitations share a common cause: existing metrics overlook the semantics of different detection behaviors, which stem from the temporal relationship of detections to anomaly events---whether they occur within an anomaly, near its boundary, or far from any anomaly.
Another limitation concerns the \textit{inconsistency caused by threshold or threshold-interval selection (L4).} Many metrics depend on model- and data-specific thresholds, which makes evaluation results sensitive to threshold choices.
Even metrics applying the Area Under the Receiver Operating Characteristic curve (AUC-ROC) or the Area Under the Precision-Recall curve (AUC-PR) still implicitly suffer from method-dependent operating threshold intervals, masking meaningful differences between models (see \autoref{subsec:Inconsistency Caused by Threshold Selection} for details).

To address L1--L3, we propose \textit{Detection Quality Evaluation (DQE)}, a novel metric for TSAD. DQE introduces a \textit{partitioning strategy} that divides a time series into local regions based on individual anomaly events, ensuring that each event is evaluated in its relevant temporal context. Each local region is further partitioned into subregions corresponding to three semantic roles of detection behavior: ground-truth (GT) detections, near-miss detections, and false alarms.
Then, by considering detections collectively rather than in isolation, detections within each subregion are grouped to evaluate overall detection quality,
capturing (i) the ability to correctly detect each anomaly event;
(ii) near-miss detection quality in terms of responsiveness, proximity, and redundancy; and (iii) false alarm characteristics, including overall burden and randomness.
Finally, DQE integrates local detection quality across
the entire threshold spectrum, providing a consistent, threshold-free assessment, thereby addressing L4.

Summarizing our main contributions:
\begin{itemize}[leftmargin=*]
    \item We systematically analyze the limitations of existing TSAD metrics, showing that they fail to capture key semantics of detection behaviors.
    \item We introduce DQE, a novel metric that evaluates detection quality across three semantic roles: GT detection, near-miss detection, and false alarms.
    \item We propose \textit{local detection event groups}, based on a novel partitioning strategy, to assess detection qualities at the group level, enabling a finer-grained evaluation of semantically distinct detection behaviors and improving interpretability.
    \item We reveal issues with AUC-ROC or AUC-PR-based metrics, and perform evaluation across the full range of thresholds, eliminating inconsistencies caused by threshold or threshold-interval selection.
    \item Extensive experiments on synthetic and real-world data demonstrate that DQE provides more comprehensive, discriminative, interpretable, and reliable evaluations than existing popular metrics.
\end{itemize}

The rest of this paper is organized as follows:
Section~\ref{sec:Limitation Analysis} discusses the limitations of existing metrics;
Section~\ref{sec:PROPOSED METRIC} presents the proposed DQE metric;
Section~\ref{sec:EXPERIMENT_AND_RESULTS} reports experimental results;
Section~\ref{sec:BACKGROUND AND RELATED WORK} reviews related work;
and Section~\ref{sec:conclusion} concludes the paper.

\begin{table*}[htbp]
    \caption{Comparison of metric inconsistency in evaluating near-miss detections.}
    \label{tab:inconsistency_in_near_section}
    \begin{subtable}[t]{\textwidth}
        \label{tab:inconsistency_in_near_section_data}
        \resizebox{\textwidth}{!}{
            \begin{tblr}{
                colspec = {ccccccccccccc},
                rowspec = {Q[m]Q[m]Q[m]Q[m]Q[m]Q[m]Q[m]},
                hline{1,Z} = {1pt},
                hline{3} = {3-Z}{0.75pt},
                hline{2} = {3-5}{0.5pt,leftpos = -1, rightpos = -1, endpos},
                hline{2} = {6-Z}{0.5pt,leftpos = -1, rightpos = -1, endpos},
                cell{1}{1} = {r=2}{},
                cell{1}{2} = {r=2}{},
                cell{1}{3} = {c=3}{},
                cell{1}{6} = {c=8}{},
                cell{1}{2-Z} = {font=\huge\bfseries},
                cell{2}{2-Z} = {font=\huge\bfseries},
                cell{1-Z}{1} = {font=\fontsize{17pt}{0pt}\selectfont\bfseries},
                cell{3-Z}{3-Z} = {font=\Huge},
                cell{6}{3} = {font=\Huge\bfseries},
                cell{5}{4-5} = {font=\Huge\bfseries},
                cell{3}{6-Z} = {font=\Huge\bfseries},
                cell{3-Z}{1} = {cmd=\raisebox{0.4cm}},
                cell{3-Z}{3-Z} = {cmd=\raisebox{0.4cm}},
            }

GT & \includegraphics{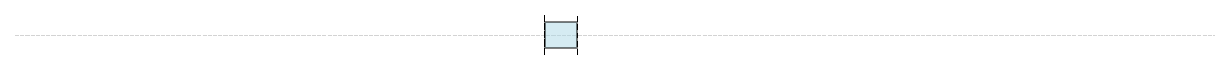} & w/ issue & VUS-PR$^{*}$ & PATE$^{*}$ & w/o issue & AUC-ROC$^{\textcolor{cyan6}{*}}$ & AUC-PR$^{\textcolor{cyan6}{*}}$ & PA-K$^{\textcolor{purple4}{*}}$ & RF$^{\star}$ & eTaF$^{\star}$ & AF$^{\star}$ & DQE$^{\star}$ \\

GT & \includegraphics{figures/single_prediction_figures/proximity_inconsistency_gt} & VUS-ROC & VUS-PR & PATE & Original-F & AUC-ROC & AUC-PR & PA-K & RF & eTaF & AF & DQE (ours) \\

P1 & \includegraphics{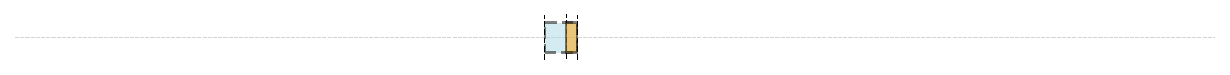} & 0.67 & 0.37 & 0.70 & 0.50 & 0.67 & 0.35 & 1.00 & 0.64 & 0.80 & 0.99 & 1.00 \\
P2 & \includegraphics{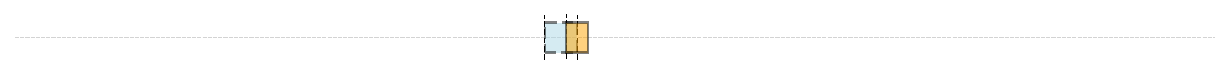} & 0.76 & 0.48 & 0.72 & 0.40 & 0.66 & 0.18 & 0.86 & 0.48 & 0.71 & 0.98 & 0.98 \\
P3 & \includegraphics{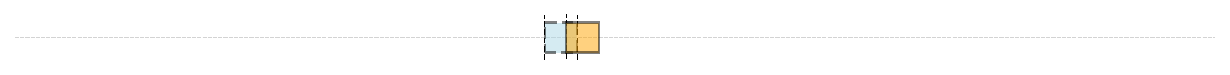} & 0.80 & 0.51 & 0.73 & 0.33 & 0.66 & 0.13 & 0.75 & 0.39 & 0.00 & 0.98 & 0.95 \\
P4 & \includegraphics{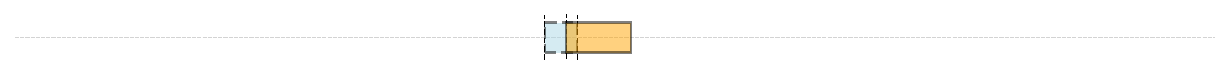} & 0.83 & 0.39 & 0.70 & 0.22 & 0.64 & 0.07 & 0.55 & 0.25 & 0.00 & 0.96 & 0.88 \\
P5 & \includegraphics{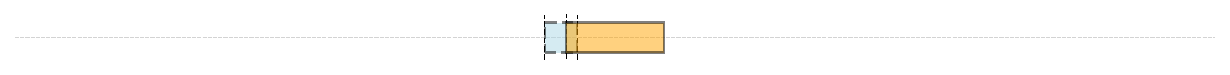} & 0.82 & 0.27 & 0.65 & 0.17 & 0.63 & 0.06 & 0.43 & 0.18 & 0.00 & 0.95 & 0.82 \\

            \end{tblr}
        }
    \end{subtable}
\end{table*}

\begin{table*}[htbp]
    \caption{Inconsistency of the AF metric in evaluating near-miss detections.}
    \label{tab:af_problem}
    \begin{subtable}[t]{\textwidth}
        \label{tab:af_problem_data}
        \resizebox{\textwidth}{!}{
            \begin{tblr}{
                colspec = {ccccccccccccc},
                rowspec = {Q[m]Q[m]Q[m]Q[m]Q[m]Q[m]Q[m]},
                hline{1,Z} = {1pt},
                hline{3} = {3-Z}{0.75pt},
                hline{2} = {3}{0.5pt,leftpos = -1, rightpos = -1, endpos},
                hline{2} = {4-Z}{0.5pt,leftpos = -1, rightpos = -1, endpos},
                cell{1}{1} = {r=2}{},
                cell{1}{2} = {r=2}{},
                cell{1}{3} = {c=1}{},
                cell{1}{4} = {c=10}{},
                cell{1}{2-Z} = {font=\huge\bfseries},
                cell{2}{2-Z} = {font=\huge\bfseries},
                cell{1-Z}{1} = {font=\fontsize{17pt}{0pt}\selectfont\bfseries},
                cell{3-Z}{3-Z} = {font=\Huge},
                cell{3}{3} = {font=\Huge\bfseries},
                cell{4}{4-Z} = {font=\Huge\bfseries},
                cell{3-Z}{1} = {cmd=\raisebox{0.4cm}},
                cell{3-Z}{3-Z} = {cmd=\raisebox{0.4cm}},
            }

GT & \includegraphics{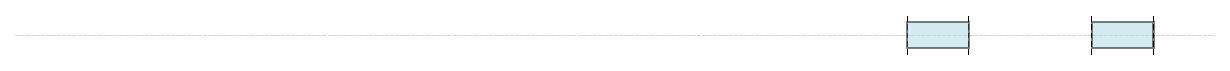} & w/ issue & w/o issue & AUC-ROC & AUC-PR & PA-K & VUS-ROC & VUS-PR & PATE & RF & eTaF & DQE \\

GT & \includegraphics{figures/single_prediction_figures/af_problem_gt} & AF & Original-F & AUC-ROC & AUC-PR & PA-K & VUS-ROC & VUS-PR & PATE & RF & eTaF & DQE (ours) \\

P1 & \includegraphics{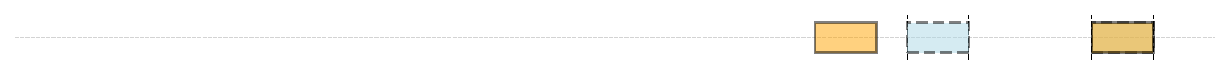} & 0.93 & 0.50 & 0.72 & 0.30 & 0.50 & 0.60 & 0.23 & 0.54 & 0.50 & 0.50 & 0.64 \\
P2 & \includegraphics{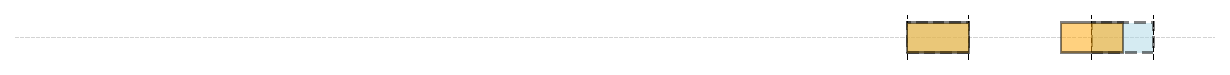} & 0.90 & 0.75 & 0.86 & 0.59 & 0.89 & 0.89 & 0.71 & 0.77 & 0.77 & 0.88 & 0.96 \\

            \end{tblr}
        }
    \end{subtable}
\end{table*}

\begin{table*}[htbp]
    \caption{Comparison of insufficient penalties for frequent false alarms across metrics.}
    \label{tab:false_alarm_only_num}
    \begin{subtable}[t]{\textwidth}
        \label{tab:false_alarm_only_num_data}
        \resizebox{\textwidth}{!}{
            \begin{tblr}{
                colspec = {ccccccccccccc},
                rowspec = {Q[m]Q[m]Q[m]Q[m]Q[m]Q[m]Q[m]},
                hline{1,Z} = {1pt},
                hline{3} = {3-Z}{0.75pt},
                hline{2} = {3-10}{0.5pt,leftpos = -1, rightpos = -1, endpos},
                hline{2} = {11-Z}{0.5pt,leftpos = -1, rightpos = -1, endpos},
                cell{1}{1} = {r=2}{},
                cell{1}{2} = {r=2}{},
                cell{1}{3} = {c=8}{},
                cell{1}{11} = {c=3}{},
                cell{1}{2-Z} = {font=\huge\bfseries},
                cell{2}{2-Z} = {font=\huge\bfseries},
                cell{1-Z}{1} = {font=\fontsize{17pt}{0pt}\selectfont\bfseries},
                cell{3-Z}{3-Z} = {font=\Huge},
                cell{3}{11-Z} = {font=\Huge\bfseries},
                cell{3-Z}{1} = {cmd=\raisebox{0.4cm}},
                cell{3-Z}{3-Z} = {cmd=\raisebox{0.4cm}},
            }

GT & \includegraphics{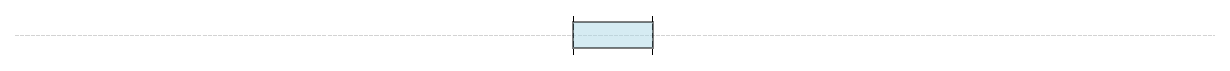} & w/ issue & VUS-PR$^{*}$ & PATE$^{*}$ & w/o issue & AUC-ROC$^{\textcolor{cyan6}{*}}$ & AUC-PR$^{\textcolor{cyan6}{*}}$ & PA-K$^{\textcolor{purple4}{*}}$ & RF$^{\star}$ & w/o issue & AF$^{\star}$ & DQE$^{\star}$ \\

GT & \includegraphics{figures/single_prediction_figures/false_alarm_only_num_gt} & Original-F & AUC-ROC & AUC-PR & PA-K & VUS-ROC & VUS-PR & PATE & AF & RF & eTaF & DQE (ours) \\

P1 & \includegraphics{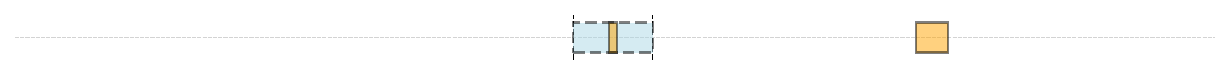} & 0.13 & 0.54 & 0.08 & 0.13 & 0.54 & 0.10 & 0.21 & 0.72 & 0.36 & 0.42 & 0.68 \\

P2 & \includegraphics{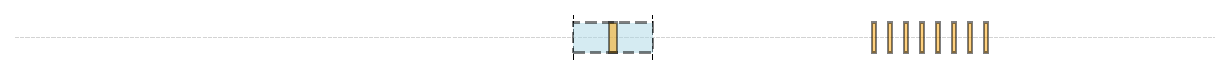} & 0.13 & 0.54 & 0.08 & 0.13 & 0.54 & 0.10 & 0.21 & 0.72 & 0.16 & 0.24 & 0.54 \\
            \end{tblr}
        }
    \end{subtable}
\end{table*}

\begin{table*}[htbp]
    \caption{Comparison of metric overvaluation for random detections.}
    \label{tab:random_case}
    \begin{subtable}[t]{\textwidth}
        \label{tab:proximity_case1}
        \resizebox{\textwidth}{!}{
            \begin{tblr}{
                colspec = {ccccccccccccc},
                rowspec = {Q[m]Q[m]Q[m]Q[m]Q[m]Q[m]Q[m]},
                hline{1,Z} = {1pt},
                hline{3} = {3-Z}{0.75pt},
                hline{2} = {3-5}{0.5pt,leftpos = -1, rightpos = -1, endpos},
                hline{2} = {6-Z}{0.5pt,leftpos = -1, rightpos = -1, endpos},
                cell{1}{1} = {r=2}{},
                cell{1}{2} = {r=2}{},
                cell{1}{3} = {c=3}{},
                cell{1}{6} = {c=8}{},
                cell{1}{2-Z} = {font=\huge\bfseries},
                cell{2}{2-Z} = {font=\huge\bfseries},
                cell{1-Z}{1} = {font=\fontsize{17pt}{0pt}\selectfont\bfseries},
                cell{3-Z}{3-Z} = {font=\Huge},
                cell{3-Z}{3-5} = {font=\Huge\bfseries},
                cell{3-Z}{1} = {cmd=\raisebox{0.4cm}},
                cell{3-Z}{3-Z} = {cmd=\raisebox{0.4cm}},
            }

GT & \includegraphics{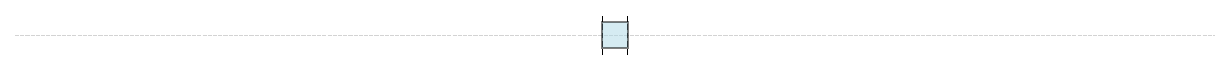} & w/ issue & VUS-ROC & AUC-ROC & w/o issue & w/o issue & PA-K & VUS-PR & RF & eTaF & Original-F & DQE \\

GT & \includegraphics{figures/single_prediction_figures/random_case_gt} & AF & VUS-ROC & AUC-ROC & PATE & Original-F & PA-K & AUC-PR & VUS-PR & RF & eTaF & DQE (ours) \\

P & \includegraphics{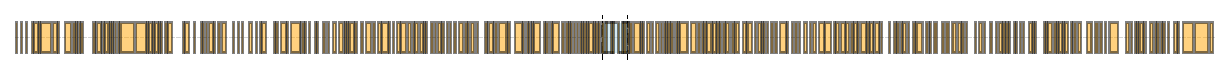} & 0.67 & 0.51 & 0.43 & 0.31 & 0.03 & 0.08 & 0.02 & 0.03 & 0.05 & 0.04 & 0.00 \\

            \end{tblr}
        }
    \end{subtable}
\end{table*}

\section{Limitation Analysis}
\label{sec:Limitation Analysis}

In this section, we systematically analyze the limitations of existing evaluation metrics for TSAD.

\subsection{L1: Bias toward Point-Level Coverage}
\label{subsec:Bias to point-level coverage proportion}

A fundamental limitation of existing metrics is their bias toward point-level coverage proportion rather than the coverage of distinct anomaly events.
As illustrated in \autoref{tab:bias_point_level_coverage}, P1 detects only a single anomaly event, whereas P2 successfully detects all anomaly events.
However, most metrics assign a higher score to P1 because it covers a larger proportion of anomalous points.
This behavior obscures missed anomaly events and may lead to misleading conclusions, which is particularly problematic in safety-critical applications such as fault diagnosis and financial fraud monitoring.

This issue is further exacerbated by the inherent imprecision of anomaly labels.
Anomaly boundaries are often context-dependent, and small variations in labeling can significantly affect point-level coverage proportion.
Moreover, many detectors produce detections only at the beginning, middle, or end of an anomaly event.
Such detections are systematically underestimated when evaluation relies heavily on point-level coverage, despite correctly indicating the positions of anomalies.

\subsection{L2: Insensitivity or Inconsistency in Near-Miss Detections}\label{subsec:Insensitivity or inconsistency evaluation in near areas of anomaly}

Detections occurring near an anomaly still carry significant value due to the temporal correlation.
However, most metrics are insensitive to proximity and assign identical scores regardless of how close a detection is to the anomaly (\autoref{tab:proximity_case}).
Even metrics that explicitly incorporate proximity, such as the VUS family and PATE, exhibit inconsistent behavior.
As shown in \autoref{tab:inconsistency_in_near_section}, their scores first increase and then decrease as detection duration extends.
Consequently, the maximum score favors less well-aligned detections (e.g., P4 for VUS-ROC, P3 for VUS-PR and PATE) over the best-aligned one (P1).
Similarly, \autoref{tab:af_problem} shows that AF assigns a higher score to a farther, non-overlapping detection (P1) than to a closer, partially overlapping one (P2),
perversely rewarding worse proximity alignment.

\subsection{L3: Inadequate Penalization of False Alarms}\label{subsec:Insufficient Penalty to False Alarms}

Existing metrics often fail to penalize false alarms adequately.
Many metrics ignore the frequency of false alarm events: as shown in \autoref{tab:false_alarm_only_num}, P1 and P2 have the same total anomaly duration, but P2 produces multiple scattered false alarms that would trigger more unnecessary interventions, yet most metrics assign identical scores.
Moreover, several metrics overestimate random detections.
As shown in \autoref{tab:random_case}, AF, VUS-ROC, and AUC-ROC assign relatively high scores to random detections, weakening their discriminative ability and masking meaningful performance differences between algorithms.
A robust evaluation metric should effectively penalize randomness and provide a sufficiently wide score range to distinguish detection quality.

\subsection{L4: Inconsistency Caused by Threshold or Threshold-Interval Selection}\label{subsec:Inconsistency Caused by Threshold Selection}

A key challenge in TSAD evaluation is the reliance on decision thresholds to convert anomaly scores into binary detections.
Common practice selects the threshold that maximizes a chosen metric for each model and each time series, introducing evaluation inconsistency: models are evaluated under different, optimized thresholds across datasets, making results highly sensitive to threshold selection.
Consequently, slight changes in the selection procedure can significantly alter results, undermining both fairness and reliability.

\begin{figure}[htbp]
  \centering
  \includegraphics[width=\columnwidth]{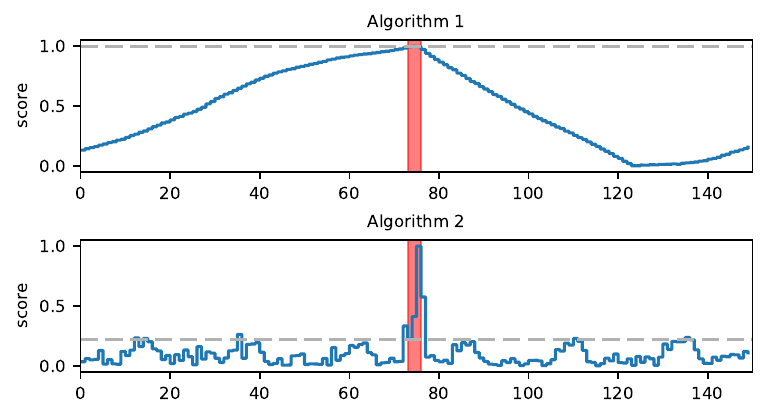}
  \caption{Output anomaly scores of various algorithms, normalized to the range [0, 1].}
  \label{fig:auc_problem_case}
\end{figure}
\begin{figure}[htbp]
    \centering
    \begin{subfigure}[t]{0.482\columnwidth}
        \includegraphics[width=\textwidth]{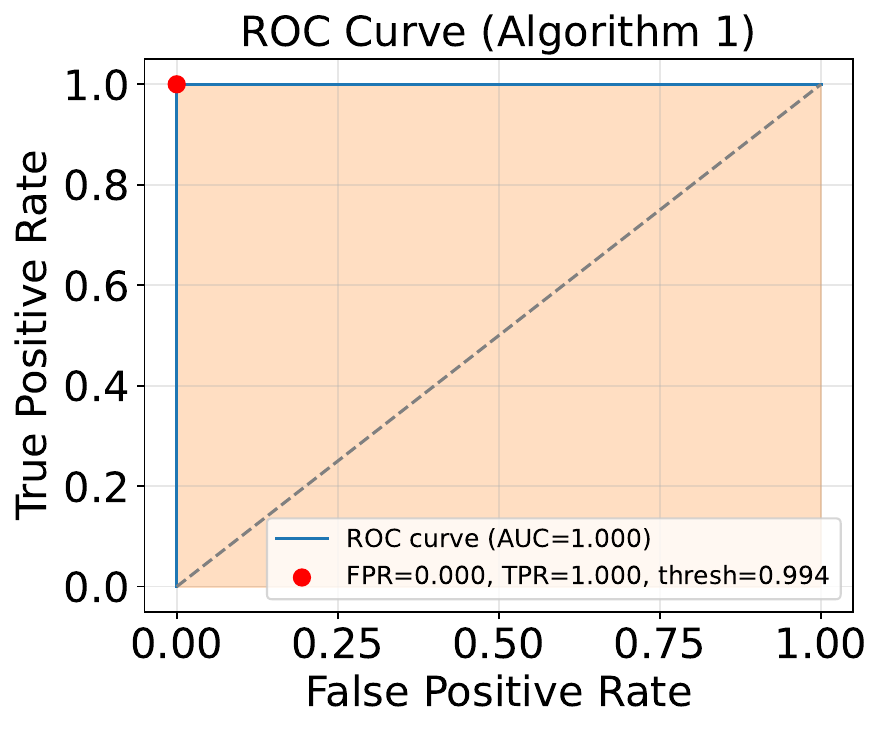}
        \caption{AUC-ROC of Algorithm 1.}
        \label{fig:auc_problem_roc_1}
    \end{subfigure}
    \hfill
    \begin{subfigure}[t]{0.482\columnwidth}
        \includegraphics[width=\columnwidth]{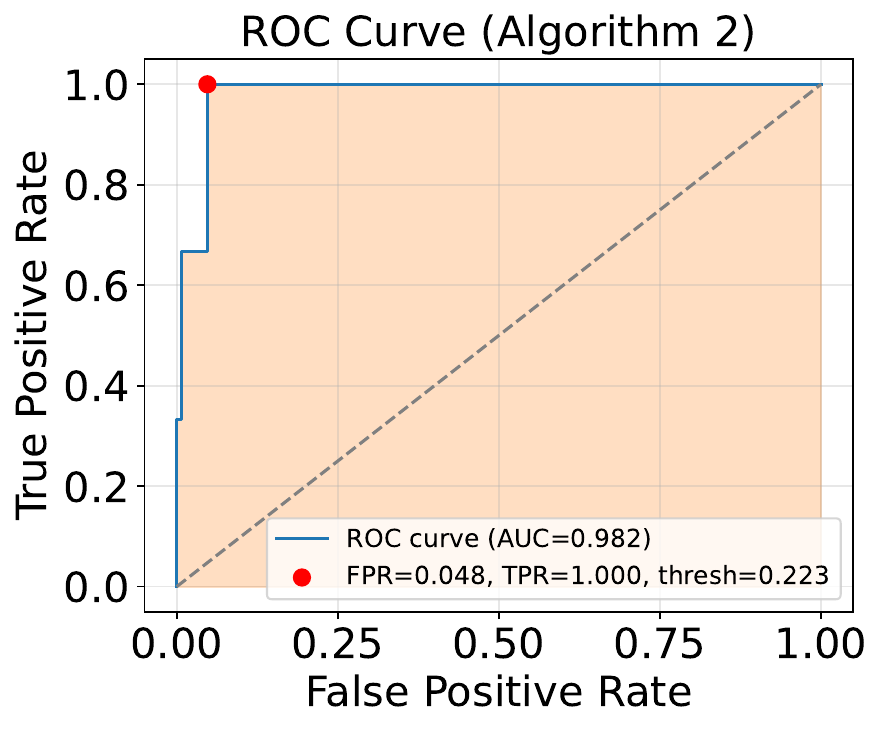}
        \caption{AUC-ROC of Algorithm 2.}
        \label{fig:Position-aware label function of Precision}
    \end{subfigure}
    \\
    \begin{subfigure}[t]{0.482\columnwidth}
        \includegraphics[width=\textwidth]{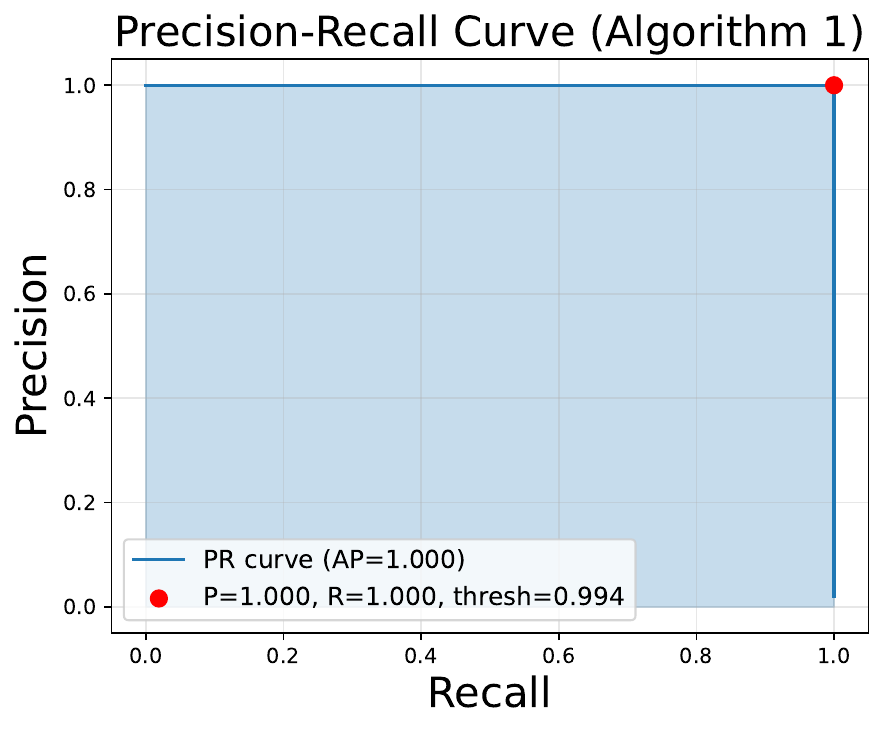}
        \caption{AUC-PR of Algorithm 1.}
        \label{fig:auc_problem_pr_1}
    \end{subfigure}
    \hfill
    \begin{subfigure}[t]{0.482\columnwidth}
        \includegraphics[width=\columnwidth]{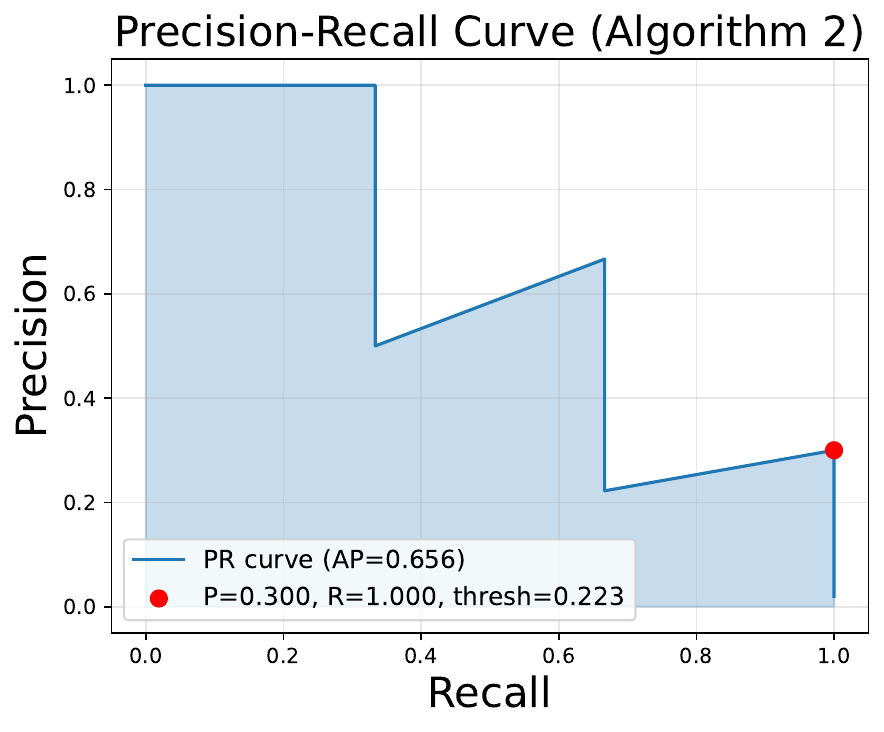}
        \caption{AUC-PR of Algorithm 2.}
        \label{fig:auc_problem_pr_2}
    \end{subfigure}
    \caption{AUC curves of two algorithms.}
    \label{fig:auc_problem_roc_pr}
\end{figure}

Even with metrics based on AUC-ROC or AUC-PR, evaluation inconsistency remains.
As shown in \autoref{fig:auc_problem_case}, Algorithm 2 produces more precise and concentrated anomaly responses with a clear separation between normal and abnormal patterns, while Algorithm 1 responds over a broader range.
However, AUC-ROC and AUC-PR (computed using Average Precision (AP), following standard practice) assign a higher score to Algorithm 1 (\autoref{fig:auc_problem_roc_pr}).
The dashed lines in \autoref{fig:auc_problem_case} indicate that Algorithm 1 has a narrow valid threshold range (1 to 0.994), whereas Algorithm 2 maintains effective performance over a much wider range (1 to 0.223).
This occurs because recall saturates at 1 over a range of thresholds, causing lower thresholds to be ignored.
As a result, AUC becomes dependent on the effective threshold interval, compromising its ability to reliably distinguish model performance.

\begin{figure}[htbp]
  \centering
  \resizebox{\columnwidth}{!}{
    \includegraphics[width=\columnwidth]{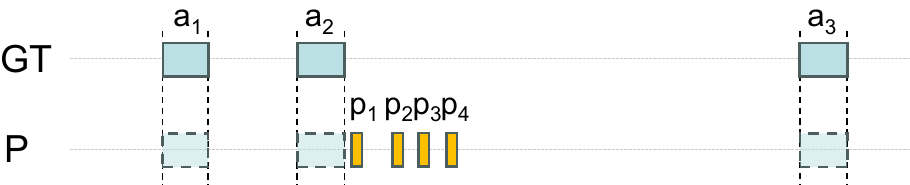}
  }
\caption{Locality and integrity of detections. $a_{1-3}$ represent anomaly events; $p_{1-4}$ represent detection events.}
  \label{fig:integrity}
\end{figure}

\begin{figure*}[htbp]
  \centering
  \includegraphics[width=\textwidth]{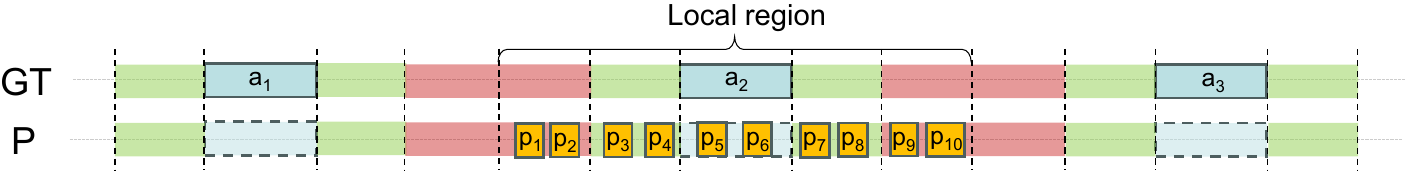}
  \caption{Partitioning strategy and local detection event groups. $a_{1-3}$ denote anomaly events; $p_{1-10}$ denote detection events. Blue, green, and red regions represent $A_{\mathtt{cap}}$, $A_{\mathtt{nm}}$, and $A_{\mathtt{fa}}$, respectively. $p_5$–$p_6$ belong to $D_{\mathtt{cap}}$; $p_3$–$p_4$ and $p_7$–$p_8$ belong to $D_{\mathtt{nm}}$; $p_1$–$p_2$ and $p_9$–$p_{10}$ belong to $D_{\mathtt{fa}}$.}
  \label{fig:partiton_strategy_and_detection_group}
\end{figure*}

\section{PROPOSED METRIC}\label{sec:PROPOSED METRIC}

\subsection{Preliminary Concepts}\label{subsec: Preliminary Concepts}

\subsubsection{Time Series Anomaly Detection}

Time series anomaly detection aims to identify abnormal patterns within time series data.
A time series is represented as $\bm{X} = \{\bm{x}_t\}_{t=1}^T$, where $\bm{x}_t \in \mathbb{R}^{m}$, $m=1$ for univariate and $m>1$ for multivariate time series, and $T$ denotes the sequence length.
An anomaly detector produces a sequence of real-valued anomaly scores, denoted as $O = \{o_t\}_{t=1}^T$, which forms the basis for evaluation.
Since evaluation operates on the output scores, the same procedure applies to both univariate and multivariate time series.
By applying a threshold to $O$, we obtain a binary detection sequence $Y = \{y_t\}_{t=1}^T$, where $y_t \in \{0, 1\}$ and $y_t=1$ indicates an anomalous point.
A detection event is defined as a contiguous temporal interval in which $y_t=1$.

\subsection{Partitioning Strategy and Local Detection Event Group}\label{subsec:Partition Strategy and Local Detection Event Group}

\subsubsection{Partitioning Strategy}

In real-world time series, anomalies, whether single points or contiguous segments, occur as semantically independent temporal events.
As illustrated in \autoref{fig:integrity}, detection $p_1$ may be a late response to $a_2$, but its relation to the distant $a_1$ is negligible.
Therefore, detections carry context-dependent semantics associated with individual anomalies, which are ignored when all detections are evaluated globally, making attribution difficult.
To better reflect detection semantics and improve interpretability, we adopt a localized evaluation strategy that partitions the time series into regions centered on individual anomaly events, as illustrated in \autoref{fig:partiton_strategy_and_detection_group}.
The boundary between two adjacent local regions is defined as the midpoint between the end of the preceding anomaly and the start of the succeeding one.
For anomalies without neighbors (corner cases),
the local region extends to the sequence boundary.

Within a local region, detections carry different semantic meanings depending on their temporal relation to the anomaly. Detections overlapping the anomaly indicate successful capture; detections near the anomaly boundary indicate contextually relevant signals, such as early/delayed detections; and detections far from the anomaly constitute false alarms.
Accordingly, each local region is further partitioned into three functional subregions: $A_\mathtt{cap}$, covering the GT anomaly to measure capture success; $A_\mathtt{nm}$, spanning the extended range surrounding the anomaly to assess near-miss detections; and $A_\mathtt{fa}$, encompassing the remaining regions to evaluate false alarms.

\subsubsection{Local Detection Event Group}

Individual detection points or events can be misleading when considered in isolation. For example, considering only detection $p_1$ in \autoref{fig:integrity} might suggest a well-performing near miss; however, when evaluated together with $p_{2-4}$, the overall result reveals poor performance due to the additional detections introduced by $p_{2-4}$.
Therefore, we evaluate detections at the event-group level within each local subregion.
Specifically, we define a local detection event group as the set of all detection events falling within a given subregion, denoted as $\mathit{D} = \{d_j\}_{j=1}^K$, where $K$ indicates the total number of detection events in that subregion, and each $d_j$ corresponds to an individual detection event.
Accordingly, $D_{\mathtt{cap}}$, $D_{\mathtt{nm}}$, and $D_{\mathtt{fa}}$ represent local detection event groups within $A_{\mathtt{cap}}$, $A_{\mathtt{nm}}$, and $A_{\mathtt{fa}}$, respectively.
If a detection event spans multiple subregions, it is split and assigned to the corresponding subregions.

\subsection{Local Evaluation}\label{subsec:Local Evaluation}

\subsubsection{Capture of the GT Anomaly Event}
\label{subsubsec:Capture to Anomaly in GT Area}

To avoid bias induced by point-level coverage and to better align evaluation with the semantic meaning of anomaly events,
anomaly capture in $A_{\mathtt{cap}}$ is evaluated at the event level.
The capture quality score is defined as:
\begin{equation} \label{eq:S_cap}
    \resizebox{\width}{!}{ $
    S_\mathtt{cap} =
        \begin{cases}
            1, & \text{if } D_{\mathtt{cap}} \neq \emptyset, \\
            0, & \text{otherwise}.
        \end{cases}
    $ }
\end{equation}
This design reflects the true performance of a detector across diverse anomaly types, ranging from isolated points to continuous segments.

\subsubsection{Near-Miss Detections Around Anomalies}
\label{subsubsec:Proximity of Detection in Nearby Area}

For detections near an anomaly, we evaluate their quality through three complementary dimensions:
\emph{responsiveness}, \emph{proximity}, and \emph{redundancy},
capturing how quickly detections respond to the anomaly, how close they are to the anomaly, and how much redundant duration they produce.

\paragraph{Closest response time.}
For a detection event $d$ in $A_{\mathtt{nm}}$, the closest response time to the nearest anomaly boundary is defined as:
\begin{equation}
    \resizebox{\width}{!}{ $
    \eta_{\mathtt{d,nm}} =
    \begin{cases}
        A_s - d_e, & \text{if } d_e \le A_s, \\
        d_s - A_e, & \text{otherwise},
    \end{cases}
    $ }
    \label{eq:eta_d_prox}
\end{equation}
where $A_s$ and $A_e$ denote the start and end of the anomaly event, and $d_s$, $d_e$ denote the start and end of the detection event.
The closest response time of all $K$ detection events of $D_{\mathtt{nm}}$ is defined as:
\begin{equation}
    \resizebox{\width}{!}{ $
    \eta_{\mathtt{D,nm}} =
    \min_{1 \le j \le K} \eta_{\mathtt{d,nm},j},
    $ }
    \label{eq:eta_D_prox}
\end{equation}
where $\eta_{\mathtt{d,nm},j}$ is the response time of the $j$-th detection event.
This captures the closest responsiveness of detections to the anomaly.

\paragraph{Mean distance.}
The mean distance of a detection event in $A_{\mathtt{nm}}$ to the anomaly event is defined as:
\begin{equation}
    \resizebox{\width}{!}{ $
    \xi_{\mathtt{d,nm}} =
    \begin{cases}
        A_s - \frac{d_s + d_e}{2}, & \text{if } d_e \le A_s, \\
        \frac{d_s + d_e}{2} - A_e, & \text{otherwise}.
    \end{cases}
    $ }
    \label{eq:xi_d_prox}
\end{equation}
The group-level mean distance of all $K$ detection events of $D_{\mathtt{nm}}$ is computed as:
\begin{equation}
    \resizebox{\width}{!}{ $
    \xi_{\mathtt{D,nm}} =
    \frac{1}{K} \sum_{j=1}^{K} \xi_{\mathtt{d,nm},j}.
    $ }
    \label{eq:xi_D_prox}
\end{equation}
Smaller mean distances indicate detections are more concentrated around the anomaly, reflecting better proximity.

\paragraph{Total duration.}
Let $\zeta_{\mathtt{D,nm}}$ denote the total duration of all $K$ detection events of $D_{\mathtt{nm}}$:
\begin{equation}
    \resizebox{\width}{!}{ $
    \zeta_{\mathtt{D,nm}} =
    \sum_{j=1}^{K} \zeta_{\mathtt{d,nm},j}.
    \label{eq:td_D_prox}
    $ }
\end{equation}
A shorter total duration indicates fewer redundant detections and thus higher precision.

\paragraph{Near-miss quality score.}
Each attribute score is mapped to a normalized value in $[0,1]$ using a monotonically decreasing linear function.
The near-miss quality score is defined as:
\begin{equation}
    \resizebox{\width}{!}{ $
    \hat{S}_{\mathtt{nm}} =
    \left(1 - \frac{\eta_{\mathtt{D,nm}}}{L_{\mathtt{nm}}}\right)
    \left(1 - \frac{\xi_{\mathtt{D,nm}}}{L_{\mathtt{nm}}}\right)
    \left(1 - \frac{\zeta_{\mathtt{D,nm}}}{L_{\mathtt{nm}}}\right),
    $ }
    \label{eq:Score_ddl}
\end{equation}
where $L_{\mathtt{nm}}$ is a normalization factor defined as the length of $A_{\mathtt{nm}}$.
It can be determined based on expert knowledge or, by default, set to half the period $\tau$ of the time series, an intrinsic property shared across all algorithms.
The reason for setting $L_{\mathtt{nm}} = \tau/2$ is that the total near-miss region around an anomaly spans one full period, resulting in half a period on each side.
This multiplicative formulation ensures that high performance is achieved only when all three aspects are favorable, enhancing the discriminative ability for near-miss detections.
When $D_{\mathtt{nm}} = \emptyset$, $\hat{S}_{\mathtt{nm}}$ is set to 1 by default.

\subsubsection{False Alarm Detections in Distant Subregion}
\label{subsubsec:False Alarm Detection in Distant Area}

False alarms are evaluated from two aspects:
the \emph{overall burden} of false alarms and their \emph{temporal randomness}.

\paragraph{Overall burden of false alarms.}
Let $\zeta_{\mathtt{D,fa}}$ denote the total duration of detections in the distant false alarm subregion $A_{\mathtt{fa}}$.
Longer false alarm durations imply higher operational cost and wasted resources.
We therefore define a duration-based false alarm score in $[0,1]$ as:
\begin{equation}
    \resizebox{\width}{!}{ $
    \tilde{S}_{\mathtt{fa}} = 1 - \frac{\zeta_{\mathtt{D,fa}}}{L_{\mathtt{fa}}/2},
    $ }
    \label{eq:S_fa_only_duration}
\end{equation}
where $L_{\mathtt{fa}}/2$ is used as a normalization factor, corresponding to the expected duration under random detections.
When $D_{\mathtt{fa}} = \emptyset$, $\tilde{S}_{\mathtt{fa}}$ is set to 1 by default.

\paragraph{Temporal randomness of false alarms.}
Scattered false alarms cause substantial user fatigue.
Therefore, we penalize temporally dispersed false alarms.
Let $I = [-a, b]$ be the range of time difference in $A_{\mathtt{fa}}$ relative to $A_{\mathtt{nm}}$, where negative values indicate
time points preceding the anomaly.
The range is partitioned into $n = a + b$ unit-length bins.
For the $k$-th bin $bin_k$, $k \in \{1, \dots, n\}$, we define a binary occupancy indicator as:
\begin{equation}
    \resizebox{\width}{!}{ $
    z_k =
    \begin{cases}
    1, & \text{if } \exists j \in \{1, \dots, K\} \text{ s.t. } d_j \in D_\mathtt{fa} \text{ and } d_j \cap bin_k \neq \emptyset, \\
    0, & \text{otherwise}.
    \end{cases}
    $ }
\end{equation}
The occupancy probability of the $k$-th bin is defined as:
\begin{equation}
    \resizebox{\width}{!}{ $
    p_k = \frac{z_k}{\sum_{i=1}^{n} z_i}.
    $ }
\end{equation}

We quantify the randomness of false alarm detections using normalized Shannon entropy in $[0,1]$:
\begin{equation}
    \resizebox{\width}{!}{ $
    S_r = \frac{-\sum_{k=1}^{n} p_k \log_2 p_k}{\log_2 n},
    $ }
\end{equation}
and define the randomness penalty coefficient as
\begin{equation}
    \alpha = 1 - S_r.
\end{equation}

A lower $\alpha$ indicates more random and widely spread false alarms, warranting a stronger penalty.

\paragraph{False alarm quality score.}
The false alarm quality score jointly considers
the overall burden and the temporal randomness:
\begin{equation}
    \resizebox{\width}{!}{ $
    \hat{S}_{\mathtt{fa}} = \alpha \cdot \tilde{S}_{\mathtt{fa}}.
    $ }
    \label{eq:S_fa}
\end{equation}

\begin{figure*}[htbp]
    \centering
    \begin{subfigure}[t]{0.33\textwidth}
        \includegraphics[width=\textwidth]{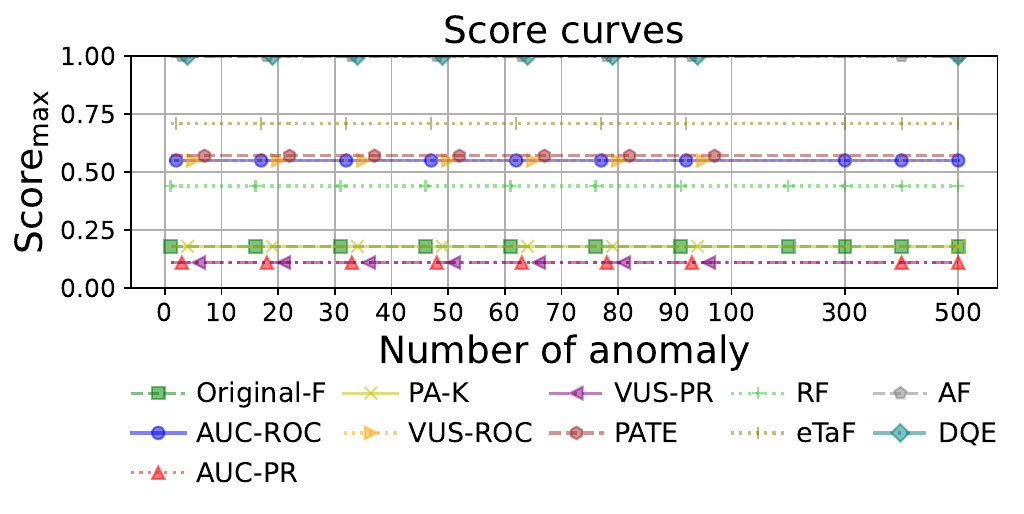}
        \vspace{-22pt}
        \caption{}
\label{fig:synthetic_exp_score_max_anomaly_num}
    \end{subfigure}
    \hfill
    \begin{subfigure}[t]{0.33\textwidth}
        \includegraphics[width=\textwidth]{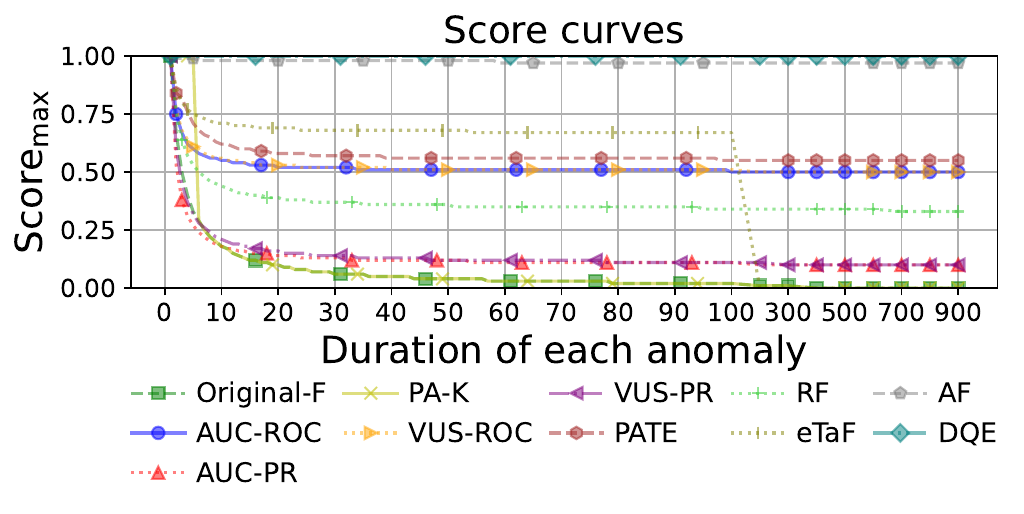}
        \vspace{-22pt}
        \caption{}
        \label{fig:synthetic_exp_score_max_anomaly_point_len}
    \end{subfigure}
        \hfill
    \begin{subfigure}[t]{0.33\textwidth}
        \includegraphics[width=\textwidth]{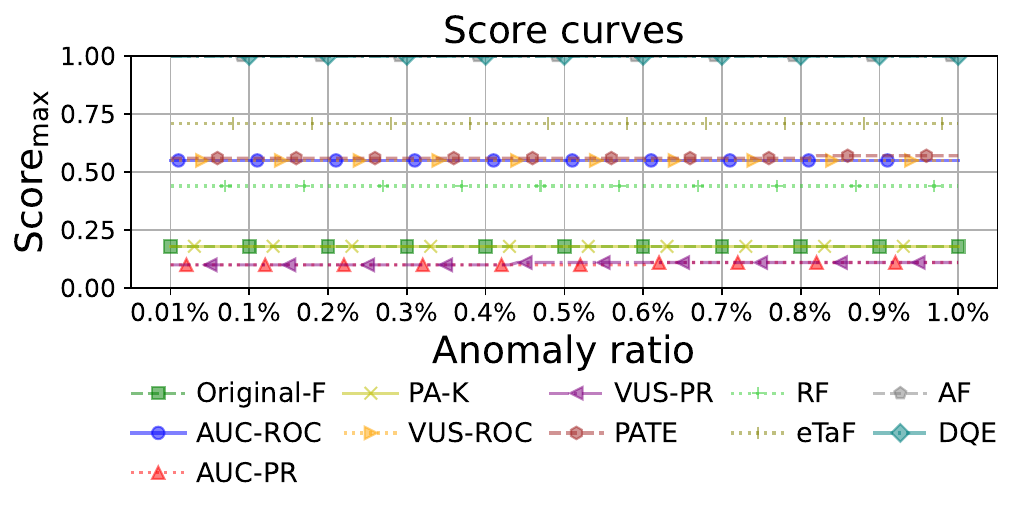}
        \vspace{-22pt}
        \caption{}
        \label{fig:synthetic_exp_score_max_anomaly_ratio}
    \end{subfigure}
    \\
    \begin{subfigure}[t]{0.33\textwidth}
        \includegraphics[width=\textwidth]{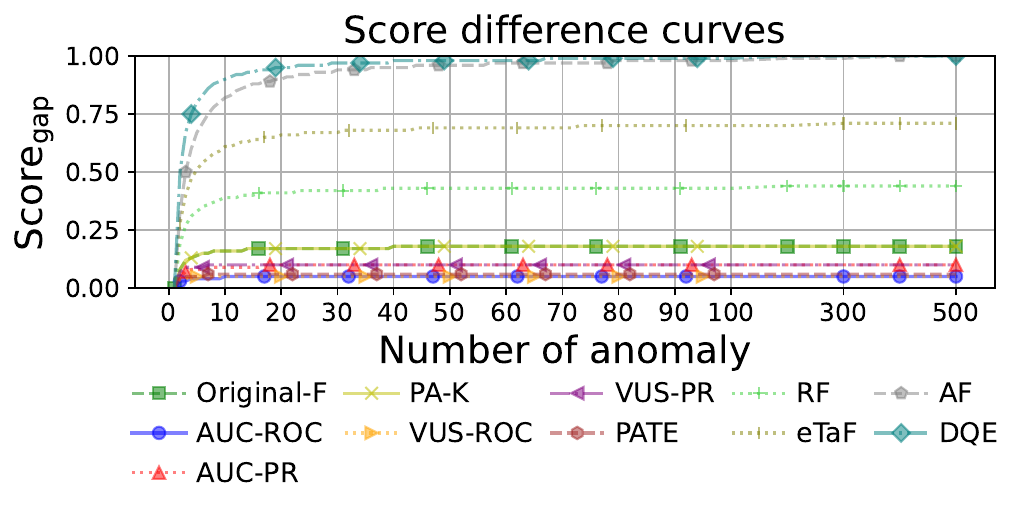}
        \vspace{-22pt}
        \caption{}
        \label{fig:synthetic_exp_score_dif_anomaly_num}
    \end{subfigure}
    \hfill
    \begin{subfigure}[t]{0.33\textwidth}
        \includegraphics[width=\textwidth]{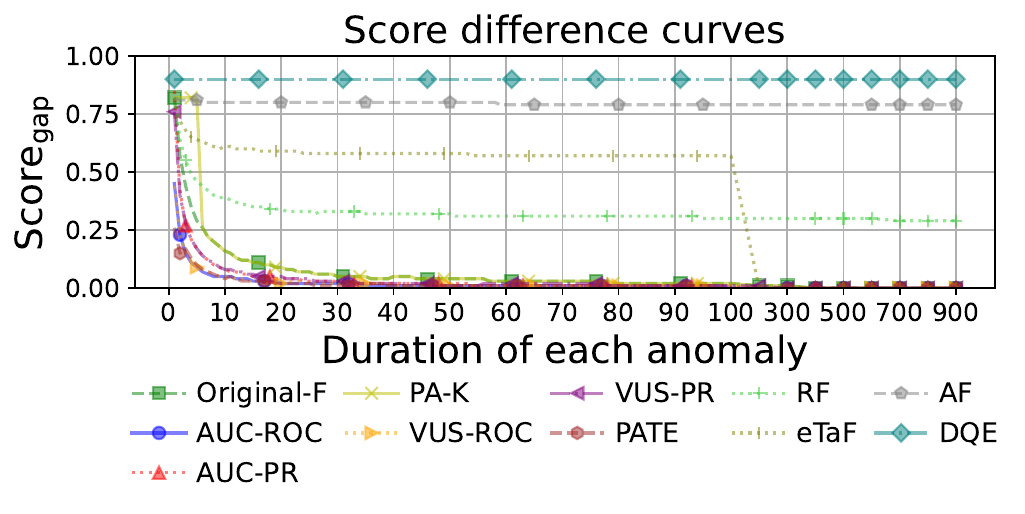}
        \vspace{-22pt}
        \caption{}
        \label{fig:synthetic_exp_score_dif_anomaly_point_len}
    \end{subfigure}
    \hfill
    \begin{subfigure}[t]{0.33\textwidth}
        \includegraphics[width=\textwidth]{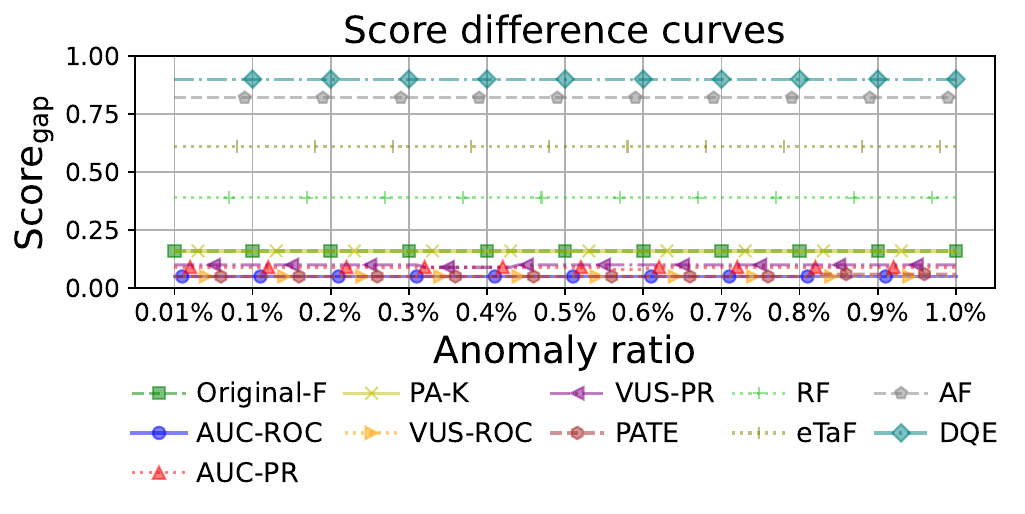}
        \vspace{-22pt}
        \caption{}
        \label{fig:synthetic_exp_score_dif_anomaly_ratio}
    \end{subfigure}
\caption{Score and score-difference curves of various metrics under variations in the number of anomalies (a, d), anomaly duration (b, e), and anomaly ratio (c, f).}
    \label{fig:synthetic_exp_score_max_dif}
\end{figure*}

\subsection{Context-Aware Score Adjustment}
\label{subsec:Context-Aware Score Adjustment}

The near-miss and false alarm quality scores are meaningful only in the context of successful anomaly capture.
For example, having no detections in areas near an anomaly is desirable only if the anomaly itself is detected; otherwise, it provides no useful information.
We therefore introduce a \emph{context-aware adjustment} function based on overall detection performance within a local region to ensure that
scores reflect meaningful detection behavior within each local detection context.

\paragraph{Adjustment for near-miss quality score.}
For the local anomaly event, the near-miss quality score is adjusted as:
\begin{equation}
    \resizebox{\width}{!}{ $
    S_{\mathtt{nm}} =
    \begin{cases}
        0, & \text{if } D_{\mathtt{cap}}= \emptyset \text{ and } D_{\mathtt{nm}}= \emptyset, \\
        0, & \text{if } D_{\mathtt{cap}} \neq \emptyset \text{ and } D_{\mathtt{fa}} \neq \emptyset \text{ and } D_{\mathtt{nm}}= \emptyset, \\
        \hat{S}_{\mathtt{nm}}, & \text{otherwise}.
    \end{cases}
    $ }
    \label{eq:adjusted_fq_near_score}
\end{equation}
This adjustment suppresses misleading rewards when the anomaly is missed or when severe false alarms exist.
\autoref{asec:Scenarios of Context-Aware Score Adjustment} provides illustrative scenarios where near-miss quality scores are suppressed due to failed anomaly capture or the presence of false alarms.

\paragraph{Adjustment for false alarm quality score.}
Similarly, when detections in both $A_{\mathtt{cap}}$ and $A_{\mathtt{nm}}$ are empty, an empty $D_{\mathtt{fa}}$ indicates that the model produces no meaningful response to the local anomaly,
In this case, the false alarm quality score is adjusted as:
\begin{equation}
    \resizebox{\width}{!}{ $
    S_{\mathtt{fa}} =
    \begin{cases}
        0, & \text{if } D_{\mathtt{cap}}= \emptyset \text{ and } D_{\mathtt{nm}}= \emptyset \text{ and } D_{\mathtt{fa}}= \emptyset, \\
        \hat{S}_{\mathtt{fa}}, & \text{otherwise}.
    \end{cases}
    $ }
    \label{eq:adjusted_proximity_score}
\end{equation}
This design ensures that false alarm evaluation is activated only when the model exhibits meaningful detection behavior.
\autoref{asec:Scenarios of Context-Aware Score Adjustment} illustrates scenarios where false alarm quality scores are suppressed due to the absence of effective detection.

\subsection{The Final DQE}
\label{subsec:The Final DQE}

Having defined fine-grained scores for different detection qualities, we now integrate them into a unified evaluation metric.
DQE jointly evaluates valuable anomaly-associated detections, including both anomaly captures and near-miss detections, as well as spurious detections, represented by false alarms.
For each anomaly event, the local DQE score at a single threshold is defined as a balance between the quality of valuable detections and that of spurious detections:
\begin{equation}
\resizebox{\width}{!}{ $
    SDQE_{\mathtt{local}} = \sqrt{\frac{S_{\mathtt{cap}} + S_{\mathtt{nm}}}{2} \cdot S_{\mathtt{fa}}}.
$ }
\end{equation}
where the square root preserves the scale of the original scores while maintaining balanced contributions.

The threshold-free local DQE is derived by averaging detection qualities across all $M$ thresholds spanning the full score spectrum:
\begin{equation}
    \resizebox{\width}{!}{ $
    DQE_{\mathtt{local}} = \frac{1}{M} \sum_{i=1}^{M} SDQE_{\mathtt{local}, i}.
    $ }
\end{equation}
By incorporating performance across all thresholds, DQE avoids inconsistencies caused by threshold or threshold-interval selection.

For evaluation across multiple anomaly events, the local DQE scores are aggregated by averaging over all $N$ anomaly events:
\begin{equation}
\resizebox{\width}{!}{ $
    DQE = \frac{1}{N} \sum_{j=1}^{N} DQE_{\mathtt{local}, j}.
$ }
\end{equation}

\section{Experiments and Results}\label{sec:EXPERIMENT_AND_RESULTS}

\subsection{Experimental Setup}\label{subsec:Experimental_Setup}

\subsubsection{Baseline Metrics}\label{subsubsec:Baseline_Metrics}

To ensure comprehensive and fair comparisons, we benchmark DQE against ten widely used evaluation metrics:
Original-F, AUC-ROC, AUC-PR, PA-K, VUS-ROC, VUS-PR, PATE, RF, eTaF, and AF.

\subsubsection{Datasets}\label{subsubsec:Datasets}
Experiments are conducted on both synthetic and real-world data.

\paragraph{Synthetic Data.} Following established methodology~\cite{ghorbani2024pate}, our synthetic experiments operate directly on binary detection sequences.
Specifically, we construct binary detection sequences to create challenging evaluation scenarios that expose issues in existing metrics.
This approach allows for a focused and unbiased evaluation of the metrics' intrinsic properties by eliminating confounding factors introduced by threshold-selection mechanisms.

\paragraph{Real-World Data.}
A major challenge in TSAD evaluation lies in the quality and reliability of GT labels in public datasets, which can undermine fair and fine-grained comparisons.
We build upon the commendable curation of a recent benchmark~\cite{liu2024elephant},
which handles mislabeling issues and dataset bias, and filters labels that lack in-context data or exhibit unrealistic anomaly ratios for anomaly detection.
Following this, we select 228 high-quality time series from UCR~\cite{wu2021current} dataset and 111 from WSD~\cite{von2018anomaly} dataset.

\begin{table*}[h!]
\centering
\caption{Algorithm rankings across different metrics for the WSD case study. Counter-intuitive rankings are underlined; over- and under-evaluations are highlighted in red and cyan, respectively.}
\label{tab:094_WSD_case}
\resizebox{\textwidth}{!}{%
    \SetTblrInner{rowsep=6pt}
    \begin{tblr}{
                    hline{1,Z} = {1pt},
                    hline{2} = {0.75pt},
                    vline{2} = {2}{0.5pt,abovepos = -1},
                    vline{2} = {3-4}{0.5pt},
                    vline{2} = {Z}{0.5pt,belowpos = -1},
                    vline{9} = {2-Z}{6pt,white},
                    cell{1}{1-Z} = {c},
                    cell{1-Z}{1} = {c},
                    cell{2-Z}{2-Z} = {l},
                    cell{1}{1-Z} = {font=\Huge\bfseries},
                    cell{2-Z}{1} = {font=\Huge\bfseries},
                    cell{2-Z}{2-Z} = {font=\Huge},
                    cell{2-Z}{12} = {font=\Huge\bfseries},
                    cell{3}{2} = {cmd=\underline},
                    cell{4}{3} = {cmd=\underline},
                    cell{3}{4} = {cmd=\underline},
                    cell{3}{5} = {cmd=\underline},
                    cell{3}{7} = {cmd=\underline},
                    cell{2}{8} = {cmd=\underline},
                    cell{4}{9} = {cmd=\underline},
                    cell{3}{10} = {cmd=\underline},
                    cell{2}{3} = {cmd=\underline},
                    cell{2}{6} = {cmd=\underline},
                    cell{2}{7} = {cmd=\underline},
                    cell{2}{3} = {fg=red},
                    cell{2}{6} = {fg=red},
                    cell{2-5}{11}  = {fg=red},
                    cell{5}{2} = {fg=cyan6},
                    cell{5}{4} = {fg=cyan6},
                    cell{5}{5} = {fg=cyan6},
                    cell{4}{7} = {fg=cyan6},
                    cell{5}{8} = {fg=cyan6},
                    cell{5}{9} = {fg=cyan6},
                    cell{5}{10} = {fg=cyan6},
                    cell{2}{4} = {fg=cyan6},
                    cell{4}{4} = {fg=cyan6},
                    cell{5}{7} = {fg=cyan6},
                    cell{4}{10} = {fg=cyan6},
                }
Rank & Original-F & AUC-ROC & AUC-PR & PA-K & VUS-ROC & VUS-PR & PATE & RF & eTaF & AF & DQE (ours) \\

1 & 0.29 (CNN) & 0.99 (Sub-LOF) & 0.19 (CNN) & 0.57 (CNN) & 0.99 (Sub-LOF) & 0.52 (Sub-LOF) & 0.66 (FFT) & 0.51 (CNN) & 0.74 (CNN) & 1.00 (CNN) & 0.79 (CNN) \\
2 & 0.23 (FFT) & 0.70 (TimesNet) & 0.14 (FFT) & 0.42 (FFT) & 0.91 (TimesNet) & 0.30 (FFT) & 0.34 (Sub-LOF) & 0.49 (Sub-LOF) & 0.38 (FFT) & 0.99 (TimesNet) & 0.54 (Sub-LOF) \\
3 & 0.20 (Sub-LOF) & 0.60 (FFT) & 0.08 (Sub-LOF) & 0.21 (Sub-LOF) & 0.85 (CNN) & 0.05 (TimesNet) & 0.24 (CNN) & 0.44 (FFT) & 0.14 (Sub-LOF) & 0.99 (Sub-LOF) & 0.46 (TimesNet) \\
4 & 0.01 (TimesNet) & 0.55 (CNN) & 0.00 (TimesNet) & 0.01 (TimesNet) & 0.72 (FFT) & 0.04 (CNN) & 0.02 (TimesNet) & 0.00 (TimesNet) & 0.00 (TimesNet) & 0.81 (FFT) & 0.34 (FFT) \\

    \end{tblr}
}
\end{table*}

\begin{table*}[h!]
\centering
\caption{Algorithm rankings across different metrics for the UCR case study. Over-evaluations are highlighted in red.}
\label{tab:452_ucr_case}
\resizebox{\textwidth}{!}{
    \SetTblrInner{rowsep=6pt}
    \begin{tblr}{
                    hline{1,Z} = {1pt},
                    hline{2} = {0.75pt},
                    vline{2} = {2}{0.5pt,abovepos = -1},
                    vline{2} = {3-4}{0.5pt},
                    vline{2} = {Z}{0.5pt,belowpos = -1},
                    vline{9} = {2-Z}{6pt,white},
                    cell{1}{1-Z} = {c},
                    cell{1-Z}{1} = {c},
                    cell{2-Z}{2-Z} = {l},
                    cell{1}{1-Z} = {font=\Huge\bfseries},
                    cell{2-Z}{1} = {font=\Huge\bfseries},
                    cell{2-Z}{2-Z} = {font=\Huge},
                    cell{2-Z}{12} = {font=\Huge\bfseries},
                    cell{2}{2} = {cmd=\underline},
                    cell{2}{3} = {cmd=\underline},
                    cell{2}{4} = {cmd=\underline},
                    cell{2}{6} = {cmd=\underline},
                    cell{2}{7} = {cmd=\underline},
                    cell{2}{8} = {cmd=\underline},
                    cell{2}{9} = {cmd=\underline},
                    cell{2}{3} = {fg=red},
                    cell{2}{6} = {fg=red},
                    cell{2-3}{5}  = {fg=red},
                    cell{2-3}{11}  = {fg=red},
                }

Rank & Original-F & AUC-ROC & AUC-PR & PA-K & VUS-ROC & VUS-PR & PATE & RF & eTaF & AF & DQE (ours) \\

1 & 0.52 (KMeansAD) & 0.99 (KMeansAD) & 0.51 (KMeansAD) & 1.00 (CNN) & 0.99 (KMeansAD) & 0.15 (KMeansAD) & 0.55 (KMeansAD) & 0.63 (KMeansAD) & 0.76 (CNN) & 1.00 (CNN) & 0.91 (CNN) \\
2 & 0.51 (CNN) & 0.69 (CNN) & 0.42 (CNN) & 0.98 (KMeansAD) & 0.83 (CNN) & 0.11 (CNN) & 0.44 (CNN) & 0.45 (CNN) & 0.75 (KMeansAD) & 1.00 (KMeansAD) & 0.50 (KMeansAD) \\

    \end{tblr}
}
\end{table*}

\begin{figure}[htbp]
  \centering
  \includegraphics[width=\columnwidth]{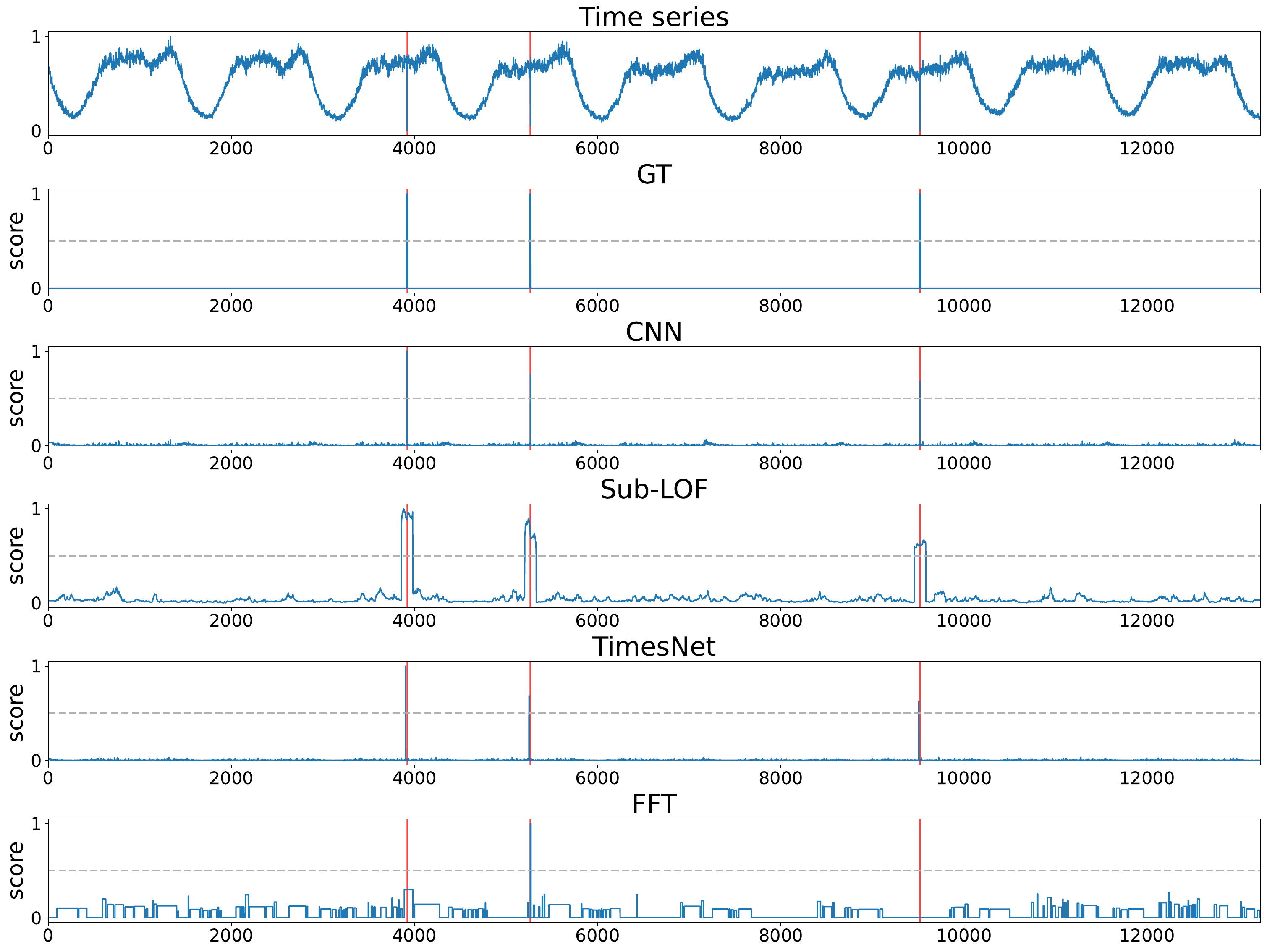}
  \caption{
    Normalized output scores for the WSD dataset case.
  }
  \label{fig:094_wsd_case}
\end{figure}

\begin{figure}[htbp]
  \centering
  \includegraphics[width=\columnwidth]{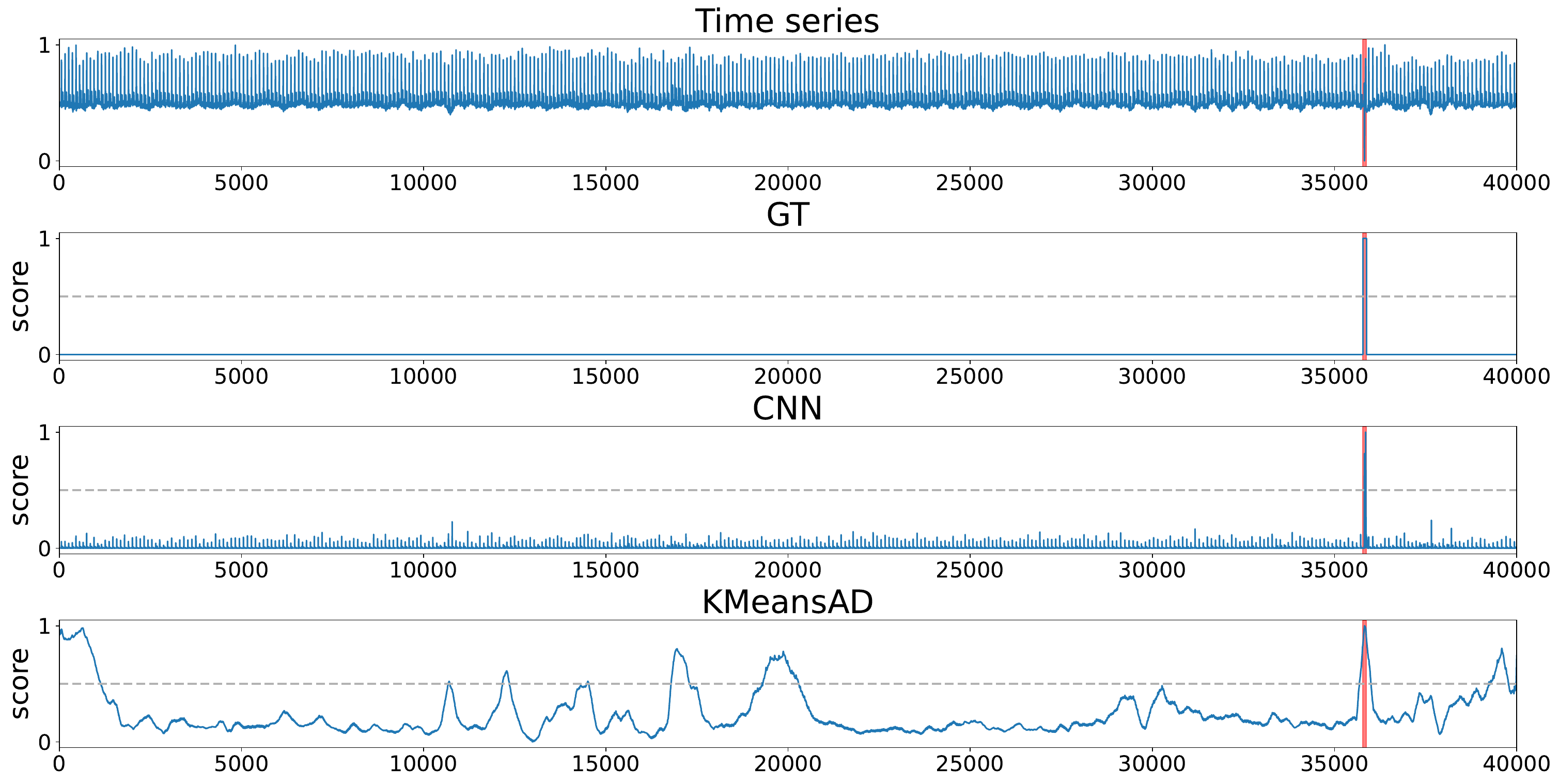}
  \caption{
    Normalized output scores for the UCR dataset case.
  }
  \label{fig:452_ucr_case}
\end{figure}

\subsubsection{Anomaly Detection Algorithms}\label{subsubsec:Algorithms}

To generate diverse detection outputs, we choose a broad spectrum of algorithms from public benchmarks~\cite{liu2024elephant, wenig2022timeeval}, ranging from high-performing to moderate methods. Selected algorithms include KMeansAD (adaptive window)~\cite{yairi2001fault}, TimesNet~\cite{wu2022timesnet}, CNN~\cite{munir2018deepant}, Sub-LOF~\cite{breunig2000lof}, and FFT~\cite{wenig2022timeeval}.

\subsection{Experiments on Synthetic Data}\label{subsec:Synthetic_Experiments}

\subsubsection{Superior Assessment of Anomaly Capture}\label{subsubsec:Superior Assessment of anomaly's capture}

DQE mitigates point-level coverage bias. As shown in \autoref{tab:bias_point_level_coverage}, while most metrics favor P1 by over-counting detected points, DQE assigns higher scores to P2 by prioritizing coverage of independent anomaly events.

To further analyze the effect of this bias, we conduct controlled synthetic experiments by varying one factor at a time: the number of anomalies, anomaly length, and anomaly ratio.
These experiments evaluate the stability and discriminative ability of event-level anomaly detection assessment under different variation scenarios.
For each setting, we report two scores: (1) $score_{\mathtt{max}}$, obtained when all anomaly events are minimally detected (i.e., detecting at least one anomalous timestamp in each anomaly event); and (2) $score_{\mathtt{gap}}$, the difference between $score_{\mathtt{max}}$ and $score_{\mathtt{min}}$, where $score_{\mathtt{min}}$ is achieved by detecting the minimum number of anomaly events (i.e., detecting at least one anomalous timestamp in only one anomaly event).
A larger $score_{\mathtt{gap}}$ indicates superior event-level discriminability.

\paragraph{Effect of the Number of Anomalies.}
Fixing anomaly length to 10 and anomaly ratio to 0.01, we vary the number of anomalies. As shown in \autoref{fig:synthetic_exp_score_max_anomaly_num} and \autoref{fig:synthetic_exp_score_dif_anomaly_num}, while $score_{\mathtt{max}}$ remains stable across metrics, DQE consistently achieves the largest $score_{\mathtt{gap}}$, indicating superior event-level discriminability under varying numbers of anomalies.

\paragraph{Effect of Anomaly Length.}
Fixing the number of anomalies to 10 and anomaly ratio to 0.1, we vary anomaly length.
When the anomaly length reaches 20 (common in public datasets), as illustrated in \autoref{fig:synthetic_exp_score_max_anomaly_point_len} and \autoref{fig:synthetic_exp_score_dif_anomaly_point_len}, $score_{\mathtt{max}}$ rapidly drops for most metrics (e.g., Original-F, AUC-PR, and VUS-PR), even when all anomalies are detected, except for AF and DQE. Further, when examining the $score_{\mathtt{gap}}$, many metrics such as Original-F, VUS-PR, VUS-ROC, and PATE, shrink to near zero, losing event-level discriminability.
Although RF and AF avoid collapse, their score gaps are unstable and highly sensitive to anomaly length.
In contrast, DQE maintains the largest and most stable score gap.

\paragraph{Effect of Anomaly Ratio.}
Fixing the number of anomalies to 10 and anomaly length to 10, we vary anomaly ratio. As illustrated in \autoref{fig:synthetic_exp_score_max_anomaly_ratio} and \autoref{fig:synthetic_exp_score_dif_anomaly_ratio}, while most metrics remain relatively stable on $score_{\mathtt{max}}$ and $score_{\mathtt{gap}}$, DQE produces the largest score gap, confirming its superior discriminative capability.

\subsubsection{Sensitivity and Consistency of Near-Miss Detections}\label{subsubsec:Sensitivity and Consistency Assessment of nearby detections}

DQE provides a fine-grained and consistent assessment of near-miss detections.
It achieves the largest score difference among P1--P4 in \autoref{tab:proximity_case} and shows a monotonic decrease in score as detections extend farther from the anomaly among P1--P5 in \autoref{tab:inconsistency_in_near_section}.
Moreover, in \autoref{tab:af_problem}, DQE favors P2 over P1, correctly assessing the proximity of detections.

\subsubsection{Penalty to False Alarm Detections}\label{subsubsec:Sufficient Penalty to False alarm detections}

DQE effectively penalizes false alarm detections. As shown in \autoref{tab:false_alarm_only_num}, its scores decrease as the number of false alarms increases. Furthermore, in \autoref{tab:random_case}, by penalizing random detections, DQE preserves more scoring space for truly effective algorithms.

\subsection{Experiments on Real-World Datasets}\label{subsec:Real_Data_Evaluation}

As algorithm performance varies across anomalies, average results alone may not adequately reflect the characteristics of the metric.
We thus conduct case studies with visualization as recommended in ~\cite{wu2021current}, while average results are reported in the \autoref{asec:Experimental results of Aggregate Performance Rankings}.

\subsubsection{Case Analysis on the WSD Dataset}\label{subsubsec:Case Analysis on WSD Dataset}

\autoref{fig:094_wsd_case} shows that CNN detects almost all anomaly events, Sub-LOF detects all but with wider ranges, and FFT almost misses two events.

Existing metrics often produce counterintuitive rankings. As illustrated in \autoref{tab:094_WSD_case}, FFT, despite missing most anomaly events, is ranked higher than CNN or Sub-LOF by the majority of metrics due to bias toward point-level coverage (PATE even ranks it as best). AUC-ROC and VUS-ROC favor Sub-LOF for its wide detections, overlooking a finer assessment of near-miss detections.

Score assignment is also problematic.
AUC-ROC, VUS-ROC, and AF grant near-perfect scores (0.99) to imperfect Sub-LOF. Moreover, AF assigns uniformly high scores to all algorithms, including 0.81 to FFT, overlooking its poor event-level coverage.
Conversely, AUC-PR undervalues well-performing CNN and Sub-LOF, and VUS-PR assigns CNN only 0.04, violating intuitive analysis.

Near-miss detection is another issue.
TimesNet fails to detect the anomaly events exactly, yet produces signals very close to the GT anomalies. Most metrics, however, assign near-zero scores, failing to reflect this valuable near-miss detection behavior.

In contrast, DQE produces intuitive rankings and scores, penalizing wide or missed event-level detections while rewarding valuable near-miss detections.

\subsubsection{Case Analysis on the UCR Dataset}\label{subsubsec:Case Analysis on UCR Dataset}

\autoref{fig:452_ucr_case} shows both CNN and KMeansAD detect anomalies, while KMeansAD produces many false alarms. As shown in \autoref{tab:452_ucr_case}, most existing metrics fail to penalize excessive false alarms or give small score differences, inadvertently encouraging false alarms. In contrast, DQE assigns reasonable rankings and penalizes false alarms appropriately.

A key advantage of DQE lies in its ability to provide component-level interpretability for each individual anomaly event. The detailed component-level analysis, which demonstrates DQE's strong interpretability, is presented in the \autoref{asec:Component-Level Analysis}.

\begin{figure}[htbp]
  \centering
  \includegraphics[width=\columnwidth]{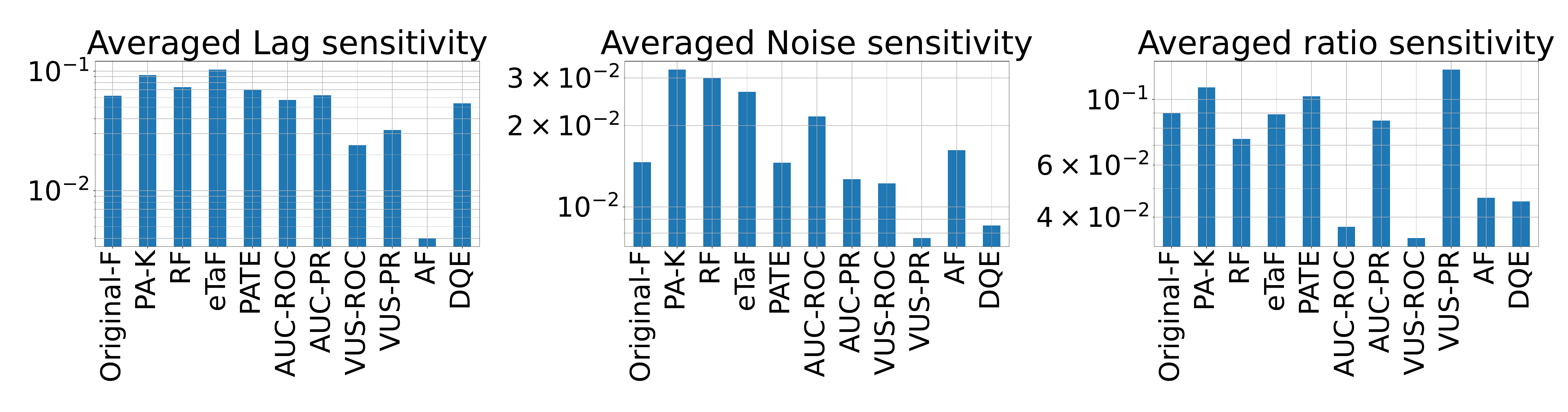}
  \caption{
    Robustness of metrics under varying lag, noise and anomaly ratios on all datasets.}
  \label{fig:robust_exp_all_dataset}
\end{figure}

\subsection{Robustness Analysis}\label{subsec:Robustness Analysis}

To assess the robustness of evaluation metrics, we vary the lag $l$, noise $n$, and anomaly ratio $r$, computing the average standard deviation across
the 11 metrics.
The perturbation ranges are defined as follows:
(a) $l \in [-0.25 * \tau, 0.25 * \tau]$, where $\tau$ denotes the time series period;
(b) $n \in [-0.05 * \Delta ST, 0.05 * \Delta ST]$, where $ST$ denotes the algorithm's output score sequence normalized to the range [0,1] and $\Delta ST = \max(ST)-\min(ST)$;
(c) $r \in [0.01, 0.2]$.
The overall robustness results across all datasets are shown in \autoref{fig:robust_exp_all_dataset}, with per-dataset results provided in the \autoref{asec:Robustness results of each dataset}.

For lag sensitivity, AF exhibits the lowest average standard deviation; however, as discussed earlier, its lack of false alarm penalties leads to inflated scores even for random detections. VUS-ROC behaves similarly.
When false alarm penalties are taken into account, VUS-PR achieves the highest robustness, with DQE closely following and outperforming most of the remaining metrics.
Under noise perturbations, both VUS-PR and DQE maintain strong robustness.
Under anomaly ratio variations, DQE performs best when accounting for false alarm penalties, while VUS-PR performs worst, highlighting its sensitivity to varying anomaly ratios, a phenomenon that is prevalent in public benchmark datasets

Overall, DQE demonstrates balanced robustness across lag, noise, and anomaly ratio while preserving reliable evaluation.

\section{RELATED WORK}\label{sec:BACKGROUND AND RELATED WORK}

\paragraph{Point-wise metrics.}
Point-wise metrics assess detections at individual time points and aggregate them to produce a final score.
Classic point-wise metrics include F1-score, AUC-ROC, and AUC-PR.
Xu et al.~\cite{xu2018unsupervised} assume that detecting a single point within a consecutive anomalous region is sufficient, and accordingly apply a point-adjustment strategy to the F1-score.
Kim et al. \cite{kim2022towards} propose point adjusted metrics at $K\%$ (PA-K), which only adjust detections when the proportion of detected true labels reaches $K\%$.
Paparrizos et al. \cite{paparrizos2022volume} note that inaccurate GT label boundaries are inevitable and introduce VUS-ROC and VUS-PR with continuous buffer regions to enhance the robustness of evaluation.
Ghorbani et al. \cite{ghorbani2024pate} propose PATE, which categorizes points and assigns consecutive weights to them based on TP, FP, and false negative (FN) contributions.

\paragraph{Segment-wise metrics.}
Segment-wise metrics evaluate detections over contiguous anomaly intervals, treating each segment as a single unit.
Based on detection and anomaly segments, range precision/recall \cite{tatbul2018precision} and their harmonic mean, the range F-score (RF), were proposed to measure the proportion of coverage in anomalies and detections, respectively.
Similar to RF, Hwang et al. \cite{hwang2019time} consider both how many anomalies are detected and how precisely each anomaly is detected, and propose the time series aware F-score (TaF) that filters out cases with a low proportion of correctly detected anomaly and detection segments using a manually set threshold $\theta$.
Further, Hwang et al. \cite{hwang2022you} observe that existing metrics overrate imprecise or insufficient cases and improve TaF by introducing the enhanced time-series aware F-score (eTaF).
Considering detection proximity, Huet et al. \cite{huet2022local} propose the affiliation-based F-score (AF), which measures the mutual proximity between anomaly and detection segments.

\section{Conclusion}
\label{sec:conclusion}

In this paper, we systematically examine the limitations of existing TSAD evaluation metrics and propose DQE, a novel metric grounded in detection semantics, to address these limitations.
Extensive experiments show that DQE provides stable and robust evaluation with strong discriminability, as well as reliable and interpretable anomaly-level assessments.

Despite these advancements, DQE opens a promising direction for future research.
The range of the near-miss area, i.e., how far a detection can deviate from the GT anomaly while still being considered a near miss, is set to half the period of the time series in this work.
However, determining an appropriate and generalizable range for near-miss areas, which may vary across application scenarios, remains a challenging problem that we plan to further investigate in future work.

\bibliographystyle{ACM-Reference-Format}
\bibliography{bibliography}

@inproceedings{xu2018unsupervised,
  title={Unsupervised anomaly detection via variational auto-encoder for seasonal kpis in web applications},
  author={Xu, Haowen and Chen, Wenxiao and Zhao, Nengwen and Li, Zeyan and Bu, Jiahao and Li, Zhihan and Liu, Ying and Zhao, Youjian and Pei, Dan and Feng, Yang and others},
  booktitle={Proceedings of the 2018 world wide web conference},
  pages={187--196},
  year={2018}
}

@inproceedings{kim2022towards,
  title={Towards a rigorous evaluation of time-series anomaly detection},
  author={Kim, Siwon and Choi, Kukjin and Choi, Hyun-Soo and Lee, Byunghan and Yoon, Sungroh},
  booktitle={Proceedings of the AAAI Conference on Artificial Intelligence},
  volume={36},
  number={7},
  pages={7194--7201},
  year={2022}
}

@article{tatbul2018precision,
  title={Precision and recall for time series},
  author={Tatbul, Nesime and Lee, Tae Jun and Zdonik, Stan and Alam, Mejbah and Gottschlich, Justin},
  journal={Advances in neural information processing systems},
  volume={31},
  year={2018}
}

@inproceedings{hwang2019time,
  title={Time-series aware precision and recall for anomaly detection: considering variety of detection result and addressing ambiguous labeling},
  author={Hwang, Won-Seok and Yun, Jeong-Han and Kim, Jonguk and Kim, Hyoung Chun},
  booktitle={Proceedings of the 28th ACM International Conference on Information and Knowledge Management},
  pages={2241--2244},
  year={2019}
}

@inproceedings{hwang2022you,
  title={Do you know existing accuracy metrics overrate time-series anomaly detections?},
  author={Hwang, Won-Seok and Yun, Jeong-Han and Kim, Jonguk and Min, Byung Gil},
  booktitle={Proceedings of the 37th ACM/SIGAPP Symposium on Applied Computing},
  pages={403--412},
  year={2022}
}

@inproceedings{huet2022local,
  title={Local evaluation of time series anomaly detection algorithms},
  author={Huet, Alexis and Navarro, Jose Manuel and Rossi, Dario},
  booktitle={Proceedings of the 28th ACM SIGKDD Conference on Knowledge Discovery and Data Mining},
  pages={635--645},
  year={2022}
}

@article{paparrizos2022volume,
  title={Volume under the surface: a new accuracy evaluation measure for time-series anomaly detection},
  author={Paparrizos, John and Boniol, Paul and Palpanas, Themis and Tsay, Ruey S and Elmore, Aaron and Franklin, Michael J},
  journal={Proceedings of the VLDB Endowment},
  volume={15},
  number={11},
  pages={2774--2787},
  year={2022},
  publisher={VLDB Endowment}
}

@inproceedings{ghorbani2024pate,
  title={Pate: Proximity-aware time series anomaly evaluation},
  author={Ghorbani, Ramin and Reinders, Marcel JT and Tax, David MJ},
  booktitle={Proceedings of the 30th ACM SIGKDD Conference on Knowledge Discovery and Data Mining},
  pages={872--883},
  year={2024}
}

@article{song2023anomaly,
  title={Anomaly VAE-transformer: A deep learning approach for anomaly detection in decentralized finance},
  author={Song, Ahyun and Seo, Euiseong and Kim, Heeyoul},
  journal={IEEE Access},
  volume={11},
  pages={98115--98131},
  year={2023},
  publisher={IEEE}
}

@inproceedings{bustamante2024financial,
  title={Financial Fraud Detection Through the Application of Machine Learning Techniques with an Anomaly-Based Approach},
  author={Bustamante Molano, Luisa Ximena and Hern{\'a}ndez Aros, Ludivia and Guti{\'e}rrez Portela, Fernando},
  booktitle={Iberoamerican Workshop on Human-Computer Interaction},
  pages={159--172},
  year={2024},
  organization={Springer}
}

@inproceedings{nakamura2025cybercscope,
  title={CyberCScope: Mining Skewed Tensor Streams and Online Anomaly Detection in Cybersecurity Systems},
  author={Nakamura, Kota and Kawabata, Koki and Tanaka, Shungo and Matsubara, Yasuko and Sakurai, Yasushi},
  booktitle={Companion Proceedings of the ACM on Web Conference 2025},
  pages={1214--1218},
  year={2025}
}

@article{rookard2024unsupervised,
  title={Unsupervised machine learning for cybersecurity anomaly detection in traditional and Software-Defined networking environments},
  author={Rookard, Curtis and Khojandi, Anahita},
  journal={IEEE Transactions on Network and Service Management},
  year={2024},
  publisher={IEEE}
}

@article{gao2025dynamic,
  title={Dynamic deep graph convolution with enhanced transformer networks for time series anomaly detection in IoT},
  author={Gao, Rong and Chen, Zhiwei and Wu, Xinyun and Yu, Yonghong and Zhang, Li},
  journal={Cluster Computing},
  volume={28},
  number={1},
  pages={15},
  year={2025},
  publisher={Springer}
}

@article{chen2025privacy,
  title={Privacy-preserving lightweight time-series anomaly detection for resource-limited Industrial IoT edge devices},
  author={Chen, Lei and Xu, Yepeng and Li, Ming and Hu, Bowen and Guo, Haomiao and Liu, Zhaohua},
  journal={IEEE Transactions on Industrial Informatics},
  year={2025},
  publisher={IEEE}
}

@inproceedings{kaufman2025time,
  title={Time-Series Anomaly Detection of Mozi Malware in IoT Devices Using Arima and Local Outlier Factor},
  author={Kaufman, T{\"u}nde and Katona, Jozsef},
  booktitle={2025 IEEE 19th International Symposium on Applied Computational Intelligence and Informatics (SACI)},
  pages={000559--000564},
  year={2025},
  organization={IEEE}
}

@inproceedings{zhong2025multi,
  title={Multi-resolution decomposable diffusion model for non-stationary time series anomaly detection},
  author={Zhong, Guojin and Yuan, Jin and Li, Zhiyong and Chen, Long and others},
  booktitle={The Thirteenth International Conference on Learning Representations},
  year={2025}
}

@article{wu2024catch,
  title={Catch: Channel-aware multivariate time series anomaly detection via frequency patching},
  author={Wu, Xingjian and Qiu, Xiangfei and Li, Zhengyu and Wang, Yihang and Hu, Jilin and Guo, Chenjuan and Xiong, Hui and Yang, Bin},
  journal={arXiv preprint arXiv:2410.12261},
  year={2024}
}

@inproceedings{li2025tsinr,
  title={TSINR: capturing temporal continuity via implicit neural representations for time series anomaly detection},
  author={Li, Mengxuan and Liu, Ke and Chen, Hongyang and Bu, Jiajun and Wang, Hongwei and Wang, Haishuai},
  booktitle={Proceedings of the 31st ACM SIGKDD Conference on Knowledge Discovery and Data Mining V. 1},
  pages={671--682},
  year={2025}
}

@inproceedings{huang2025graph,
  title={Graph mixture of experts and memory-augmented routers for multivariate time series anomaly detection},
  author={Huang, Xiaoyu and Chen, Weidong and Hu, Bo and Mao, Zhendong},
  booktitle={Proceedings of the AAAI Conference on Artificial Intelligence},
  volume={39},
  number={16},
  pages={17476--17484},
  year={2025}
}

@inproceedings{jang2025tail,
  title={TAIL-MIL: Time-aware and instance-learnable multiple instance learning for multivariate time series anomaly detection},
  author={Jang, Jaeseok and Kwon, Hyuk-Yoon},
  booktitle={Proceedings of the AAAI Conference on Artificial Intelligence},
  volume={39},
  number={17},
  pages={17582--17589},
  year={2025}
}

@inproceedings{liu2025gcad,
  title={Gcad: Anomaly detection in multivariate time series from the perspective of granger causality},
  author={Liu, Zehao and Gao, Mengzhou and Jiao, Pengfei},
  booktitle={Proceedings of the AAAI Conference on Artificial Intelligence},
  volume={39},
  number={18},
  pages={19041--19049},
  year={2025}
}

@article{liu2024elephant,
  title={The elephant in the room: Towards a reliable time-series anomaly detection benchmark},
  author={Liu, Qinghua and Paparrizos, John},
  journal={Advances in Neural Information Processing Systems},
  volume={37},
  pages={108231--108261},
  year={2024}
}

@inproceedings{yairi2001fault,
  title={Fault detection by mining association rules from house-keeping data},
  author={Yairi, Takehisa and Kato, Yoshikiyo and Hori, Koichi},
  booktitle={Proc. of International Symposium on Artificial Intelligence, Robotics and Automation in Space},
  volume={3},
  number={9},
  year={2001}
}

@article{wu2022timesnet,
  title={Timesnet: Temporal 2d-variation modeling for general time series analysis},
  author={Wu, Haixu and Hu, Tengge and Liu, Yong and Zhou, Hang and Wang, Jianmin and Long, Mingsheng},
  journal={arXiv preprint arXiv:2210.02186},
  year={2022}
}

@article{munir2018deepant,
  title={DeepAnT: A deep learning approach for unsupervised anomaly detection in time series},
  author={Munir, Mohsin and Siddiqui, Shoaib Ahmed and Dengel, Andreas and Ahmed, Sheraz},
  journal={Ieee Access},
  volume={7},
  pages={1991--2005},
  year={2018},
  publisher={IEEE}
}

@inproceedings{breunig2000lof,
  title={LOF: identifying density-based local outliers},
  author={Breunig, Markus M and Kriegel, Hans-Peter and Ng, Raymond T and Sander, J{\"o}rg},
  booktitle={Proceedings of the 2000 ACM SIGMOD international conference on Management of data},
  pages={93--104},
  year={2000}
}

@incollection{von2018anomaly,
  title={Anomaly detection and localization for cyber-physical production systems with self-organizing maps},
  author={von Birgelen, Alexander and Niggemann, Oliver},
  booktitle={IMPROVE-Innovative Modelling Approaches for Production Systems to Raise Validatable Efficiency: Intelligent Methods for the Factory of the Future},
  pages={55--71},
  year={2018},
  publisher={Springer Berlin Heidelberg Berlin, Heidelberg}
}

@article{wenig2022timeeval,
  title={Timeeval: A benchmarking toolkit for time series anomaly detection algorithms},
  author={Wenig, Phillip and Schmidl, Sebastian and Papenbrock, Thorsten},
  journal={Proceedings of the VLDB Endowment},
  volume={15},
  number={12},
  pages={3678--3681},
  year={2022},
  publisher={VLDB Endowment}
}

@article{wu2021current,
  title={Current time series anomaly detection benchmarks are flawed and are creating the illusion of progress},
  author={Wu, Renjie and Keogh, Eamonn J},
  journal={IEEE transactions on knowledge and data engineering},
  volume={35},
  number={3},
  pages={2421--2429},
  year={2021},
  publisher={IEEE}
}

\twocolumn

\appendix

\section{Scenario Analysis of Context-Aware Score Adjustment}\label{asec:Scenarios of Context-Aware Score Adjustment}

\begin{figure}[htbp]
  \centering
    \begin{subfigure}[t]{\columnwidth}
        \includegraphics[width=\textwidth]{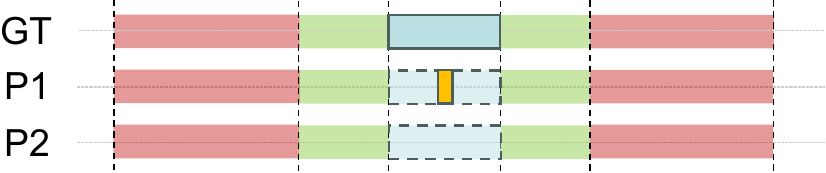}
        \caption{}
        \label{fig:adjustment_proximity_scenatio1}
    \end{subfigure}
    \\
    \begin{subfigure}[t]{\columnwidth}
        \includegraphics[width=\textwidth]{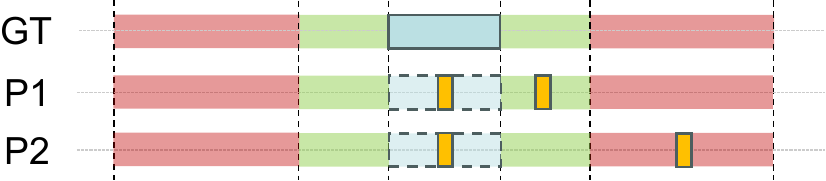}
        \caption{}
        \label{fig:adjustment_proximity_scenario2}
    \end{subfigure}
  \caption{Scenarios for score adjustment within $A_{\mathtt{nm}}$.}
  \label{fig:Score adjustment in near-miss region}
\end{figure}

We analyze context-aware adjustments for empty detections using visually illustrated scenarios.
As illustrated by P2 in \autoref{fig:adjustment_proximity_scenatio1}, if no detection occurs in $A_{\mathtt{cap}}$, indicating the absence of any detection associated with the local anomaly (unlike P1, which successfully captures the anomaly), the near-miss quality score for empty detections in $A_{\mathtt{nm}}$ is set to zero.
This adjustment reflects that near-miss detections should not be rewarded if the anomaly is entirely missed.
Another adjustment occurs when both $D_{\mathtt{cap}}$ and $D_{\mathtt{fa}}$ are non-empty, as shown in P2 of \autoref{fig:adjustment_proximity_scenario2}.
The presence of detections in $A_{\mathtt{fa}}$ indicates that detections occur outside the near-miss subregion, which is less informative than true near-miss detections in P1. Therefore, the near-miss quality score is set to 0 for empty detections in $A_{\mathtt{nm}}$.

\begin{figure}[htbp]
  \centering
  \includegraphics[width=\columnwidth]{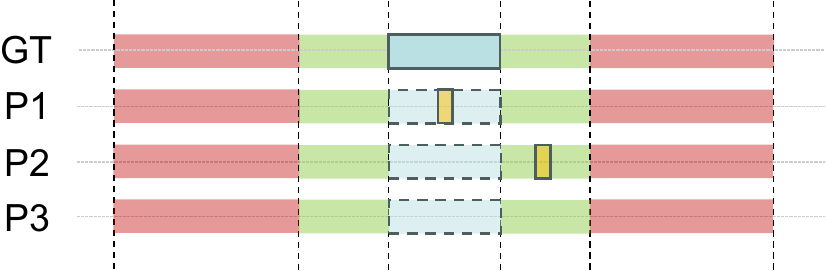}
  \caption{Scenarios for score adjustment within $A_{\mathtt{fa}}$.}
  \label{fig:Score adjustment in distant FQ region}
\end{figure}

For false alarm detections, when detections in both $A_{\mathtt{cap}}$ and $A_{\mathtt{nm}}$ are empty, as illustrated in P3 of \autoref{fig:Score adjustment in distant FQ region}, empty detections in $D_{\mathtt{fa}}$ imply that the model does nothing to the nearby anomaly, which is less favorable than valuable detections in P1 or P2.
Accordingly, the corresponding score is adjusted to zero.

\begin{table}[h!]
\centering
\caption{Analysis of DQE components for the UCR dataset case.}
\label{tab:452_UCR_case_component_analysis}
\resizebox{\columnwidth}{!}{
    \SetTblrInner{rowsep=8pt}
    \begin{tblr}{
                    hline{1,Z} = {1pt},
                    hline{2} = {0.75pt},
                    vline{2} = {2}{0.5pt,abovepos = -1},
                    vline{2} = {3-5}{0.5pt},
                    vline{2} = {Z}{0.5pt,belowpos = -1},
                    cell{1}{1-Z} = {c},
                    cell{1-Z}{1} = {c},
                    cell{2-Z}{2-Z} = {c},
                    cell{1}{1-Z} = {font=\Huge\bfseries},
                    cell{2-Z}{1} = {font=\Huge\bfseries},
                    cell{2-Z}{2-Z} = {font=\Huge},
                }
Rank & Algorithm & DQE & DQE$_\mathtt{cap}$ & DQE$_\mathtt{nm}$ & DQE$_\mathtt{fa}$ \\
1 & CNN          & 0.91  & 0.99  & 0.82  & 0.92 \\
2 & KMeansAD  & 0.50  & 1.00  & 0.13  & 0.53 \\

    \end{tblr}
}
\end{table}

\begin{table}[h!]
\centering
\caption{Algorithm rankings on the WSD dataset: comparison between sequence-level and anomaly-event-level aggregation.}
\label{tab:WSD_dataset_all_methods_order_res_sequence_level_anomaly_level_compare}
\resizebox{\columnwidth}{!}{
    \begin{tblr}{
                    hline{1,Z} = {1pt},
                    hline{2} = {0.75pt},
                    vline{2} = {2}{0.5pt,abovepos = -1},
                    vline{2} = {3-8}{0.5pt},
                    vline{2} = {Z}{0.5pt,belowpos = -1},
                    cell{1}{1-Z} = {c},
                    cell{1-Z}{1} = {c},
                    cell{2-Z}{2-Z} = {l},
                    cell{1}{1-Z} = {font=\bfseries},
                    cell{2-Z}{1} = {font=\bfseries},
                }
Rank & DQE (sequence level) & DQE (anomaly-event level) \\

1 & 0.65 (CNN) & 0.5 (CNN) \\
2 & 0.50 (Sub-LOF) & 0.41 (Sub-LOF) \\
3 & 0.42 (FFT) & 0.39 (FFT) \\
4 & 0.35 (TimesNet) & 0.25 (TimesNet) \\
5 & 0.12(KMeansAD) & 0.13 (KMeansAD) \\

    \end{tblr}
}
\end{table}

\begin{figure}[b]
  \centering
  \includegraphics[width=\columnwidth]{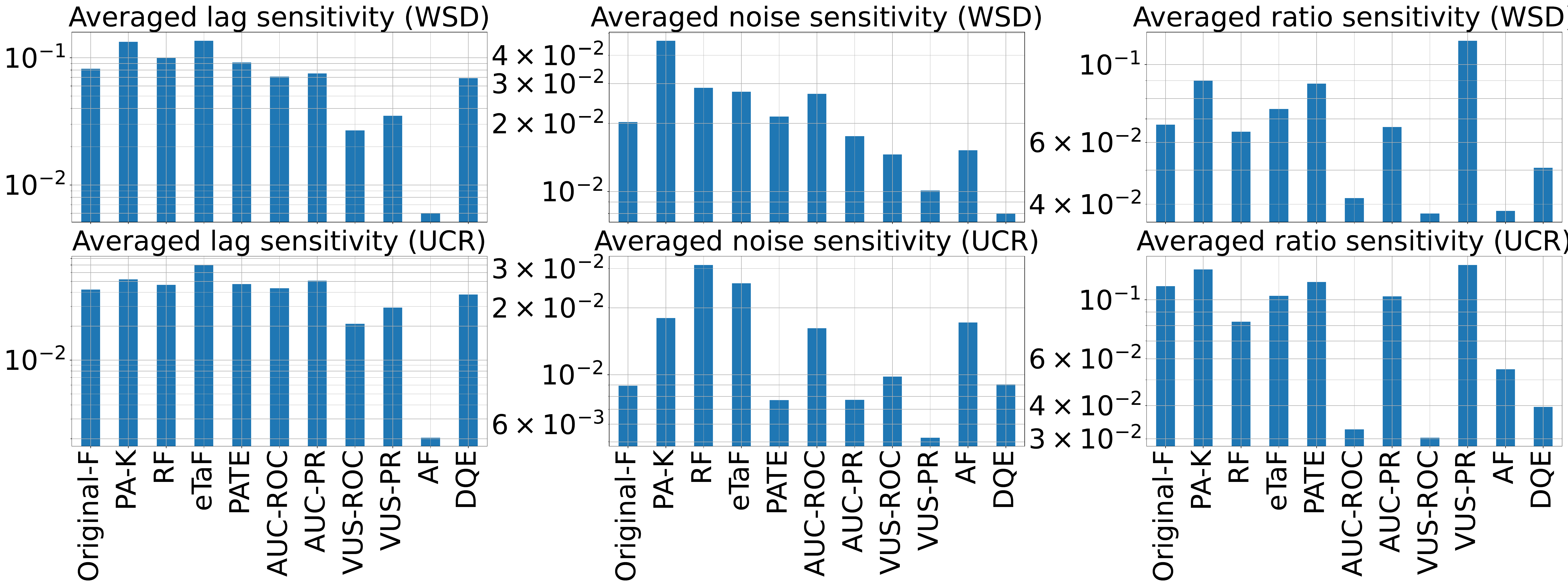}
  \caption{Average standard deviation of metric scores under varying lag, noise, and anomaly ratios on the WSD and UCR datasets.}
  \label{fig:robust_exp_each_dataset}
\end{figure}

\begin{table*}[h!]
\centering
\caption{Analysis of DQE components on the WSD dataset case for each anomaly event.}
\label{tab:094_WSD_case_component_analysis_each_anomaly}
\resizebox{\textwidth}{!}{
    \SetTblrInner{rowsep=8pt}
    \begin{tblr}{
                    hline{1,Z} = {1pt},
                    hline{3} = {0.75pt},
                    vline{2} = {3}{0.5pt,abovepos = -1},
                    vline{2} = {4-6}{0.5pt},
                    vline{2} = {Z}{0.5pt,belowpos = -1},
                    vline{3} = {3-Z}{0.5pt, dashed},
                    vline{7} = {3-Z}{0.5pt, dashed},
                    vline{11} = {3-Z}{0.5pt, dashed},
                    vline{15} = {3-Z}{0.5pt, dashed},
                    cell{1}{1-Z} = {c},
                    cell{1-Z}{1} = {c},
                    cell{2-Z}{2-Z} = {c},
                    hline{2} = {3-6}{0.5pt,leftpos = -1, rightpos = -1, endpos},
                    hline{2} = {7-10}{0.5pt,leftpos = -1, rightpos = -1, endpos},
                    hline{2} = {11-14}{0.5pt,leftpos = -1, rightpos = -1, endpos},
                    hline{2} = {15-18}{0.5pt,leftpos = -1, rightpos = -1, endpos},
                    cell{1}{1} = {r=2}{},
                    cell{1}{2} = {r=2}{},
                    cell{1}{3} = {c=4}{},
                    cell{1}{7} = {c=4}{},
                    cell{1}{11} = {c=4}{},
                    cell{1}{15} = {c=4}{},
                    cell{1}{1-Z} = {font=\Huge\bfseries},
                    cell{2-Z}{1} = {font=\Huge\bfseries},
                    cell{2-Z}{2-Z} = {font=\Huge},
                }
Rank & Algorithm & mean result & DQE$_\mathtt{cap}$ & DQE$_\mathtt{nm}$ & DQE$_\mathtt{fa}$ & first anomaly event & DQE$_\mathtt{cap}$ & DQE$_\mathtt{nm}$ & DQE$_\mathtt{fa}$ & second anomaly event & DQE$_\mathtt{cap}$ & DQE$_\mathtt{nm}$ & DQE$_\mathtt{fa}$ & third anomaly event & DQE$_\mathtt{cap}$ & DQE$_\mathtt{nm}$ & DQE$_\mathtt{fa}$ \\

Rank & Algorithm & DQE & DQE$_\mathtt{cap}$ & DQE$_\mathtt{nm}$ & DQE$_\mathtt{fa}$ & DQE$_\mathtt{local}$ & DQE$_\mathtt{local,cap}$ & DQE$_\mathtt{local,nm}$ & DQE$_\mathtt{local,fa}$ & DQE$_\mathtt{local}$ & DQE$_\mathtt{local,cap}$ & DQE$_\mathtt{local,nm}$ & DQE$_\mathtt{local,fa}$ & DQE$_\mathtt{local}$ & DQE$_\mathtt{local,cap}$ & DQE$_\mathtt{local,nm}$ & DQE$_\mathtt{local,fa}$ \\
1 & CNN       & 0.79 & 0.81 & 0.77 & 0.79
& 0.98 & 1.00 & 0.96 & 0.98
& 0.73 & 0.75 & 0.71 & 0.72
& 0.66 & 0.68 & 0.65 & 0.66 \\

2 & Sub-LOF   & 0.54 & 0.78 & 0.07 & 0.76
& 0.63 & 0.90 & 0.06 & 0.90
& 0.57 & 0.80 & 0.10 & 0.79
& 0.43 & 0.63 & 0.05 & 0.59 \\

3 & TimesNet  & 0.46 & 0.00 & 0.56 & 0.76
& 0.58 & 0.00 & 0.69 & 0.99
& 0.40 & 0.00 & 0.50 & 0.66
& 0.39 & 0.00 & 0.49 & 0.62 \\

4 & FFT       & 0.34 & 0.43 & 0.32 & 0.38
& 0.12 & 0.29 & 0.05 & 0.14
& 0.90 & 1.00 & 0.90 & 0.88
& 0.00 & 0.00 & 0.00 & 0.11 \\

    \end{tblr}
}
\end{table*}

\begin{table*}[h!]
\centering
\caption{Algorithm rankings on the WSD dataset based on average scores across different evaluation metrics.}
\label{tab:WSD_dataset_all_methods_order_res}
\resizebox{\textwidth}{!}{
    \begin{tblr}{
                    hline{1,Z} = {1pt},
                    hline{2} = {0.75pt},
                    vline{2} = {2}{0.5pt,abovepos = -1},
                    vline{2} = {3-8}{0.5pt},
                    vline{2} = {Z}{0.5pt,belowpos = -1},
                    cell{1}{1-Z} = {c},
                    cell{1-Z}{1} = {c},
                    cell{2-Z}{2-Z} = {l},
                    cell{1}{1-Z} = {font=\bfseries},
                    cell{2-Z}{1} = {font=\bfseries},
                }
Rank & Original-F & AUC-ROC & AUC-PR & PA-K & VUS-ROC & VUS-PR & PATE & RF & eTaF & AF & DQE (ours) \\

1 & 0.47 (Sub-LOF) & 0.92 (Sub-LOF) & 0.37 (Sub-LOF) & 0.71 (CNN) & 0.94 (Sub-LOF) & 0.38 (FFT) & 0.52 (FFT) & 0.52 (CNN) & 0.73 (CNN) & 0.98 (CNN) & 0.65 (CNN) \\
2 & 0.44 (FFT) & 0.82 (FFT) & 0.32 (FFT) & 0.58 (Sub-LOF) & 0.91 (TimesNet) & 0.35 (Sub-LOF) & 0.52 (Sub-LOF) & 0.51 (Sub-LOF) & 0.48 (Sub-LOF) & 0.97 (TimesNet) & 0.50 (Sub-LOF) \\
3 & 0.41 (CNN) & 0.79 (TimesNet) & 0.32 (CNN) & 0.53 (FFT) & 0.89 (FFT) & 0.21 (TimesNet) & 0.40 (CNN) & 0.45 (FFT) & 0.38 (FFT) & 0.96 (Sub-LOF) & 0.42 (FFT) \\
4 & 0.26 (TimesNet) & 0.74 (CNN) & 0.16 (TimesNet) & 0.39 (TimesNet) & 0.88 (CNN) & 0.19 (CNN) & 0.24 (TimesNet) & 0.20 (TimesNet) & 0.30 (TimesNet) & 0.91 (FFT) & 0.35 (TimesNet) \\
5 & 0.10 (KMeansAD) & 0.60 (KMeansAD) & 0.06 (KMeansAD) & 0.13 (KMeansAD) & 0.66 (KMeansAD) & 0.08 (KMeansAD) & 0.13 (KMeansAD) & 0.17 (KMeansAD) & 0.10 (KMeansAD) & 0.74 (KMeansAD) & 0.12 (KMeansAD) \\

    \end{tblr}
}
\end{table*}

\begin{table*}[h!]
\centering
\caption{Algorithm rankings on the UCR dataset based on average scores across different evaluation metrics.}
\label{tab:UCR_dataset_all_methods_order_res}
\resizebox{\textwidth}{!}{
    \begin{tblr}{
                    hline{1,Z} = {1pt},
                    hline{2} = {0.75pt},
                    vline{2} = {2}{0.5pt,abovepos = -1},
                    vline{2} = {3-8}{0.5pt},
                    vline{2} = {Z}{0.5pt,belowpos = -1},
                    cell{1}{1-Z} = {c},
                    cell{1-Z}{1} = {c},
                    cell{2-Z}{2-Z} = {l},
                    cell{1}{1-Z} = {font=\bfseries},
                    cell{2-Z}{1} = {font=\bfseries},
                }
Rank & Original-F & AUC-ROC & AUC-PR & PA-K & VUS-ROC & VUS-PR & PATE & RF & eTaF & AF & DQE (ours) \\

1 & 0.40 (Sub-LOF) & 0.86 (Sub-LOF) & 0.32 (KMeansAD) & 0.62 (Sub-LOF) & 0.89 (Sub-LOF) & 0.26 (KMeansAD) & 0.37 (Sub-LOF) & 0.40 (Sub-LOF) & 0.50 (Sub-LOF) & 0.92 (Sub-LOF) & 0.54 (Sub-LOF) \\
2 & 0.35 (KMeansAD) & 0.84 (KMeansAD) & 0.30 (Sub-LOF) & 0.45 (KMeansAD) & 0.85 (KMeansAD) & 0.25 (Sub-LOF) & 0.34 (KMeansAD) & 0.37 (KMeansAD) & 0.38 (KMeansAD) & 0.87 (CNN) & 0.42 (CNN) \\
3 & 0.13 (CNN) & 0.66 (CNN) & 0.09 (CNN) & 0.29 (CNN) & 0.74 (CNN) & 0.07 (CNN) & 0.10 (CNN) & 0.20 (CNN) & 0.31 (CNN) & 0.85 (KMeansAD) & 0.31 (TimesNet) \\
4 & 0.08 (FFT) & 0.54 (TimesNet) & 0.04 (FFT) & 0.13 (FFT) & 0.63 (TimesNet) & 0.06 (FFT) & 0.06 (FFT) & 0.11 (TimesNet) & 0.16 (TimesNet) & 0.82 (FFT) & 0.30 (FFT) \\
5 & 0.04 (TimesNet) & 0.52 (FFT) & 0.02 (TimesNet) & 0.09 (TimesNet) & 0.56 (FFT) & 0.02 (TimesNet) & 0.02 (TimesNet) & 0.11 (FFT) & 0.14 (FFT) & 0.81 (TimesNet) & 0.30 (KMeansAD) \\

    \end{tblr}
}
\end{table*}

\section{Component-Level
Analysis}\label{asec:Component-Level Analysis}

A key advantage of DQE lies in its component-level interpretability, which allows for a granular diagnosis of detection behavior for each individual anomaly event.

\autoref{tab:452_UCR_case_component_analysis} presents the component scores for each anomaly in \autoref{fig:452_ucr_case} (only a single anomaly event exists), including DQE$_{\mathtt{cap}}$, DQE$_{\mathtt{nm}}$, and DQE$_{\mathtt{fa}}$, which quantify detection quality in terms of capturing the GT anomaly, identifying near-miss detections, and avoiding false alarms, respectively.
The results of component scores indicate that DQE appropriately penalizes KMeansAD in DQE$_{\mathtt{fa}}$ (0.53) and DQE$_{\mathtt{nm}}$ (0.13) due to its excessive false alarms and wide detections near the anomaly,
consistent with the analysis in \autoref{subsubsec:Case Analysis on UCR Dataset}.

Similarly, the local component scores---DQE$_{\mathtt{local,cap}}$, DQE$_{\mathtt{local,nm}}$, and DQE$_{\mathtt{local,fa}}$---assess the corresponding detection qualities for a single anomaly event.
From the DQE$_{\mathtt{local}}$ scores in \autoref{tab:094_WSD_case_component_analysis_each_anomaly}, FFT performs poorly on the first (0.12) and third (0.00) anomalies, resulting in a low sequence-level score of 0.34, which aligns with the analysis in \autoref{subsubsec:Case Analysis on WSD Dataset}.
This component-level analysis precisely identifies which events contribute to the poor sequence-level performance, facilitating clearer attribution and enhancing interpretability of the anomaly detection results.
Additionally, even though the DQE$_{\mathtt{local,cap}}$ of each anomaly is zero on TimesNet, the good performance on near-miss detections and false alarms for each anomaly is rewarded by DQE$_{\mathtt{local,nm}}$ and DQE$_{\mathtt{local,fa}}$.
Also, Sub-LOF is penalized by DQE$_{\mathtt{nm}}$ for its wide detection ranges.

\section{Experimental Results of Aggregate Performance Rankings}\label{asec:Experimental results of Aggregate Performance Rankings}

Importantly, DQE enables evaluation at the anomaly-event level rather than solely at the sequence level. Sequence-level averaging is often biased by the imbalance in anomaly numbers across different time series (e.g., a series with 10 anomalies weighs the same as one with 1 anomaly). By assessing each anomaly event independently, DQE eliminates this bias, offering a more statistically fair and reasonable comparison of algorithmic capability.
\autoref{tab:WSD_dataset_all_methods_order_res_sequence_level_anomaly_level_compare} compares average rankings on the WSD dataset obtained at the sequence level and the anomaly-event level.

\autoref{tab:WSD_dataset_all_methods_order_res} and \autoref{tab:UCR_dataset_all_methods_order_res} report the average performance rankings of all evaluated algorithms on the WSD and UCR datasets, respectively, averaged at the sequence level across all time series.

\section{Robustness Results of Each Dataset}\label{asec:Robustness results of each dataset}

\autoref{fig:robust_exp_each_dataset} shows the average standard deviation of metric scores for each dataset.
These results align with the robustness analysis discussed in \autoref{subsec:Robustness Analysis}.

\end{document}